\documentclass{article}

\usepackage[final, nonatbib]{styles/neurips_2024}

\usepackage{amsmath}
\usepackage[utf8]{inputenc} 
\usepackage[T1]{fontenc}    
\usepackage{hyperref}       
\usepackage{url}            
\usepackage{booktabs}       
\usepackage{amsfonts}       
\usepackage{nicefrac}       
\usepackage{microtype}      
\usepackage{xcolor}         

\usepackage[super,numbers,sort&compress]{natbib} 
\usepackage{dirtytalk} 
\usepackage{graphicx} 
\usepackage{txfonts} 
\usepackage{float} 
\usepackage{caption} 
\usepackage{algorithm}
\usepackage[noend]{algpseudocode}

\algnewcommand\algorithmicinput{\textbf{Preprocessing:}}
\algnewcommand\Preprocessing{\item[\algorithmicinput]}
\algnewcommand\algorithmicequire{\textbf{Note:}}
\algnewcommand\Note{\item[\algorithmicequire]}

\algrenewcommand\algorithmicrequire{\textbf{Input:}}

\title{Upsampling DINOv2 features for unsupervised vision tasks and weakly supervised materials segmentation}

\author{
  Ronan ~Docherty\footnotemark{*} \\
  Department of Materials\\
  Department of Design Engineering\\
  Imperial College London\\
  \And
  Antonis ~Vamvakeros \\
  Department of Design Engineering\\
  Imperial College London\\
  Finden Limited\\
  \And
  Samuel J. ~Cooper \\
  Department of Design Engineering\\
  Imperial College London\\
}

\begin{document}

\maketitle

\begin{abstract}
The features of self-supervised vision transformers (ViTs) contain strong semantic and positional information relevant to downstream tasks like object localization and segmentation. Recent works combine these features with traditional methods like clustering, graph partitioning or region correlations to achieve impressive baselines without finetuning or training additional networks. We leverage upsampled features from ViT networks (\textit{e.g} DINOv2) in two workflows: in a clustering based approach for object localization and segmentation, and paired with standard classifiers in weakly supervised materials segmentation. Both show strong performance on benchmarks, especially in weakly supervised segmentation where the ViT features capture complex relationships inaccessible to classical approaches. We expect the flexibility and generalizability of these features will both speed up and strengthen materials characterization, from segmentation to property-prediction.
\end{abstract}

\section{Introduction}
\label{sec:intro}

\renewcommand{\thefootnote}{*}
\footnotetext{ronan.docherty18@imperial.ac.uk}

The rise of `foundation models' - large networks (usually transformers\cite{ATTN_IS_ALL_YOU_NEED}) trained on vast corpuses of data - has been a strong theme in machine learning for the past five years\cite{FM_SURVEY}.
Initially these were language models, such as BERT\cite{BERT}, T5\cite{T5} or the GPT\cite{GPT_1, GPT_3} series, but has also included some generative vision models like CLIP\cite{CLIP}, DALL-E\cite{DALLE} or Stable Diffusion\cite{STABLE_DIFFUSION}.
Foundation models targeting fundamental vision problems like segmentation and object detection have also been developed, notably Meta's Segment Anything Model (SAM)\cite{SAM} and the YOLO series\cite{YOLO, YOLO_REVIEW} that display strong zero- or few-shot performance. 
The analysis of micrographs is ubiquitous in scientific workflows, for example in the characterization of crystals in material science or cells in biology. Segmentation (assigning every pixel/voxel a class) is prerequisite before any such analysis can take place, be it phase quantification, defect detection or transport simulations.
Finding ways to use these models to improve materials segmentation is therefore of great interest.

One class of vision foundation model that has seen progress recently is the `feature foundation model', designed to learn \say{all-purpose visual features}\cite{DINOv2}.
These features can then be used in downstream tasks, usually by freezing the foundation model and training a small 'head' network to map from the image features to the specific objective. Examples include DINO\cite{DINO}, DINOv2\cite{DINOv2} and I-JEPA\cite{IJEPA}.
Like other foundation models, they are predominantly trained via self-supervised learning. ViTs trained in this way capture rich semantic information in their features that convolutional neural networks do not\cite{DENSE_VIT_FEATURES, DINO}.

DINO and DINOv2 produce spatialised features, albeit at a resolution limited by the model's patch size. These semantically meaningful patch-level features can then be combined with classical techniques (clustering, spectral methods \textit{etc.}) in workflows that aim to leverage these features in an unsupervised manner, \textit{i.e,} without further training\cite{VIT_UNSUPERVISED_OD_SURVEY}. 
Example tasks that can be achieved using these features unsupervised include (multi-)object detection\cite{MOST, LOST, VIT_UNSUPERVISED_NORMCUT}, saliency detection\cite{VIT_UNSUPERVISED_NORMCUT, DEEP_SPECTRAL} and (semantic) segmentation\cite{DENSE_VIT_FEATURES, DEEP_SPECTRAL}. 
These workflows require no labelled training data (unlike, say, training a head network) and are therefore attractive in domains where labelled training data is scarce and expensive, like biological, medical or materials imaging\cite{MATERIALS_DATA_DIVERSITY}.

In such domains, dense (pixel-level) features are often desirable for tasks like semantic segmentation or localised property prediction. For example, in materials science, zero-shot semantic segmentation is often achieved by training a classifier (usually a Random Forest) to map from classical image features like average local intensity, edge strength and textures to user-drawn labels\cite{WEKA, ILASTIK, NAPARI_APOC, NAPARI_FEATURE_CLASSIFIER}.
The semantic information offered by ViTs could improve tasks like segmentation where classical image features fail, but are currently limited by being patch- rather than pixel-level. Work on improving the feature resolution of these models has been done, including reducing the stride of the convolution in the model's patch embedding layer\cite{DENSE_VIT_FEATURES}, training a single forward-pass upsampling filter or fitting an implicit model for each image\cite{FEATUP}.

In this work we present three main contributions - the first is a novel single-pass method for feature upsampling that is model agnostic and works without any further training. It works by shifting the input image a certain number of pixels (less than the patch size) in each direction, computing the features with a foundation model, resizing the features back to the original image size, shifting in the opposite direction and averaging. This method is compatible with the strided approach and works for other invertible transformations like flipping, rotating \textit{etc}.  

\begin{figure}
\centering
    \includegraphics[width=1\linewidth]{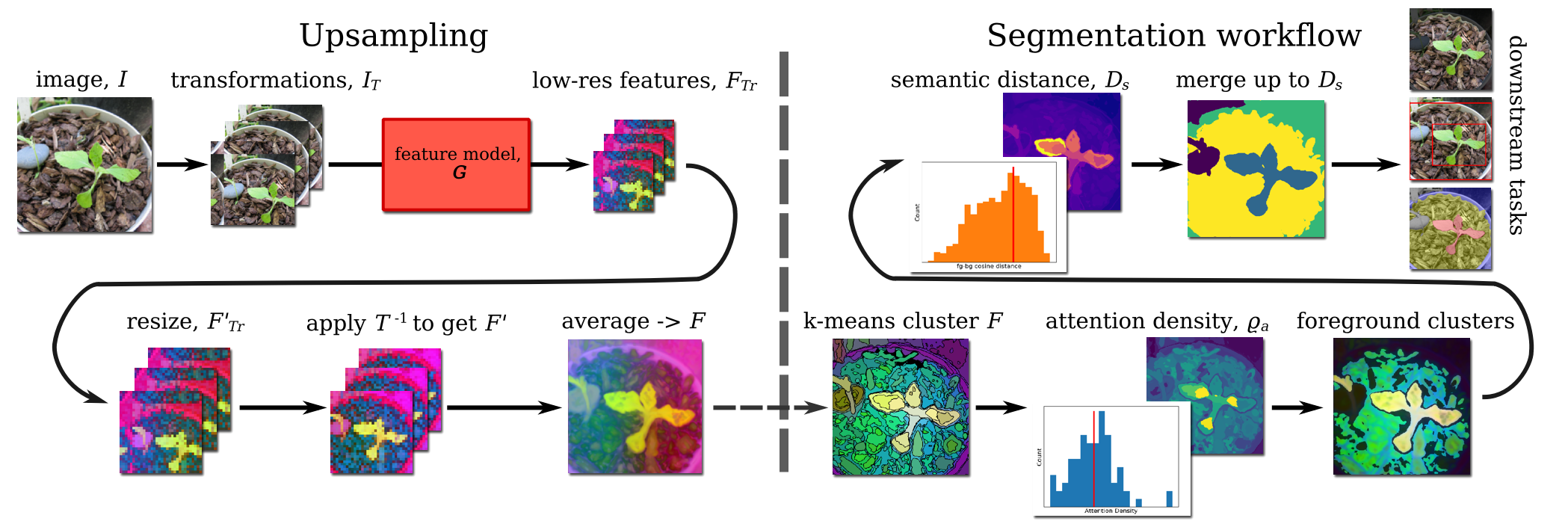}
    \caption{Diagram of our feature upsampling method (left) and its application in unsupervised downstream vision tasks like object detection, saliency detection and semantic segmentation.}
    \label{fig:explanation}
\end{figure}

The second is an unsupervised segmentation workflow based on clustering these high resolution feature maps, where the attention density can be used to create a robust foreground/background distinction, and from that a `semantic distance' between classes in the image can be estimated. Clusters are agglomeratively merged up to this distance, producing a `semantic segmentation' relative to the image (and model). This segmentation can then be processed for object detection, saliency detection or semantic segmentation. A diagram of the feature upsampling and its use in the unsupervised workflow is presented in Figure \ref{fig:explanation}.

Finally, we integrate these high resolution features into a zero-shot weakly supervised semantic segmentation app and qualitatively demonstrate its ability to handle both pixel-level and semantic distinctions when applied to a case study of cell nuclei segmentation and a series of industrially-relevant materials - namely, battery cathodes, alloys, oxide layers and organic crystals.

\section{Background}

\label{sec:background}

\subsection*{Vision transformers}
\label{sec:vit}
Vision Transformers extend the token sequence-modelling of language transformers to images by breaking an image into (non-overlapping) embedded patches with a convolutional layer with given kernel size (commonly 8, 16 or 32 pixels) and stride equal to kernel size\cite{VIT}.
These patches are given a positional encoding relative to their position in the image (\textit{i.e,} raster-order\cite{VIT} or sinusoidal\cite{FOURIER_PE}) and then treated as tokens in the transformer as normal.
It is common to append a non-spatial class token (called the \texttt{[CLS]} token) to the sequence to capture global information.

The number of patch tokens, $n$, increases quadratically as the edge length of the square image increases, or patch size decreases. 
Given the attention mechanism of transformers is $O(n^2)$\cite{ATTN_IS_ALL_YOU_NEED} this represents a sharp quartic scaling to achieve higher resolution.
It is therefore common to train these networks at a large patch size, say 16 pixels, which limits the resolution of extracted features - not usually a problem for natural images but has implications when handling micrographs of materials, which often have pixel-level features. 

\subsection*{Self-supervised feature learning}
\label{sec:ssl}
Self-supervised representation learning with transformers on large datasets has proven both popular and successful in Natural Language Processing (NLP), learning by mapping between transformations of an input text sequence - like causal token masking for GPT\cite{GPT_1} or single-token masking for BERT\cite{BERT}.

A direct analogue to this approach for ViTs is the Masked Autoencoder (MAE), which learns by reconstructing an image with around 70\% of the input patches masked; it works in different modalities and as part of wider feature learning setups\cite{MAE, AUDIO_MAE, VIDEO_MAE, IJEPA, VJEPA}.
Other approaches view self-supervised learning as a self-distillation problem, employing a student-teacher framework to learn patch-level features from either varying local/global view of an image, or distillation token or in combination with masked token modelling. Examples of these approaches are DINO\cite{DINO}, DEiT\cite{DEIT} and iBot\cite{IBOT}. The most recent feature foundation model is DINOv2 \cite{DINOv2}, which showed strong performance across various benchmarks.

\textcolor{black}{DINOv2\cite{DINOv2} iterated on the DINO formula, including iBOT loss and various training stability improvements. It trained to learn features over a curated dataset of around 140M images and showed strong performance across various benchmarks. Evaluation was done by freezing the model and training small head networks (linear probes for classification or convolutional layers for dense tasks) on the features. }

It has been shown features of supervised models are more discriminative (and therefore less flexible) than those of self-supervised models, and that self-supervised ViTs learn semantic information that comparable CNNs do not\cite{DINO, DEIT, VIT_ROBUST}. Different self-supervised training schemes result in different features: the features of MAE-trained ViTs are better at separating instances of the same object whereas DINO-trained ViTs are better at semantic separation\cite{MAE_VS_DINO_SSL}.

These feature models tend to use the same ViT architecture, varying the training data and objective. The ViT architecture tends to come in varying sizes: (S)mall, (B)ase, (L)arge, (H)uge and (G)iant, each with increasing parameter counts and hidden dimension output.
They also have varying patch size, usually 16 but sometimes 14 or 8. We will use the notation $\texttt{model-size-patch}$, \textit{i.e} DINO-S-16, in the following sections.

\subsection*{Feature upsampling}
\label{sec:feat_upsample}
Standard interpolation approaches (bilinear, bicubic, nearest-neighbour) can be used to upsample patch-level features but risks blurring or missing high-resolution details.
For ViTs, reducing the stride of the patch embedding layer (creating overlapping patches)\cite{DENSE_VIT_FEATURES} can increase the feature resolution, but this increases the number of input patches and therefore causes the time and memory scaling problem discussed in Section \ref{sec:limits}.
It can also cause blurring and at very small strides (1 or 2) cause numerical errors.

FeatUp\cite{FEATUP} presented two approaches for general feature upsampling: a learned Joint Bilateral Upsampling (JBU)\cite{JBU} filter which operates in a single forward pass and an implict model that is fitted to each image. A feature upsampler and downsampler are trained simultaneously to predict the change in features after small transformations of the input image.
The JBU filter is fast, though sometimes produces blurry feature maps and the implicit approach produces very sharp features but requires training for each image.

\subsection*{Unsupervised feature adaption: object localization and segmentation}
\label{sec:downstream}

Various methods have been proposed for applying unsupervised image features to different tasks - most leveraging DINO as their featuriser\cite{VIT_UNSUPERVISED_OD_SURVEY}.
LOST\cite{LOST} detects objects by finding patches correlated to a given `seed' patch.
MOST\cite{MOST} performs entropy-based box analysis on the patch feature correlations to detect foreground objects.

Other approaches use clustering, usually to cluster features into classes (and later superclasses) for semantic segmentation\cite{CLUSTERING_UNSUPERVISED, DENSE_VIT_FEATURES}.
Spectral methods have also been used, either decomposing images via eigenvectors of the Laplacian of the feature affinity matrix\cite{DEEP_SPECTRAL}, or graph/spectral clustering of the features\cite{DEEPCUT, MAE_VS_DINO_SSL}, or a graph-cut approach\cite{VIT_UNSUPERVISED_NORMCUT, TOKENCUT}. The memory cost of these methods scales prohibitively as the number of tokens (and therefore graph size) increases.

Each of these applications needs some heuristic for separating foreground and background classes, be it thresholding attention\cite{DINO}, number of  pixels touching the edges of the image\cite{MAE_VS_DINO_SSL} or assuming the largest class to be the background\cite{LOST}.
They also must contend with the limited resolution of the patch features, improving them by changing the model stride\cite{DENSE_VIT_FEATURES}, fusing additional colour information\cite{DEEP_SPECTRAL}, using a Bilateral Solver (BS)\cite{TOKENCUT} or Conditional Random Field (CRF)\cite{DENSE_VIT_FEATURES} or training a model on their workflow's pseudo-labels (`self-training').

\subsection*{Weakly supervised segmentation for materials}
\label{sec:weak_seg}

Weakly- or semi-supervised segmentation in computer science is a wide field with a variety of models, including CNNs, generative adversarial networks (GANs) and transformers\cite{SCRIBBLESEG, WSS_GAN, WSS_TRANSFORMER, WSS_REVIEW}.
Materials science is a field that spans a range of materials and length scales, and that often has limited access to computational hardware or large quantities of data. 
As such, cheap methods that generalise well on new data are preferable. Tools like Weka\cite{WEKA}, ilastik\cite{ILASTIK} and napari-apoc\cite{NAPARI_APOC} train a random forest to map from classical image features (Gaussian, Sobel, Hessian, Difference of Gaussians, Laplacian, \textit{etc.}, filters) to user-drawn labels.
These features are computed for every pixel over a range of scales, so are full resolution, though are limited in the complexity of the relationships they can express (relative to a neural network). 

As the featurisation only happens once and it is quick to train a random forest, users can add new labels in an active learning style\cite{AL_REVIEW}, correcting wrong or uncertain outputs to improve the segmentation\cite{WEKA, ILASTIK}. Once the classifier is trained, it can be applied to new examples without the user needing to add labels (\textit{i.e,} in an automated analysis workflow).

\section{Method}
\label{sec:method}

\subsection{Increasing feature resolution}

Our method for upsampling features is simple. Starting with a set of invertible transformations, $T = \{t_{1}, t_{2}, ...\}$ , an input image $I \in \mathbb{R}^{h \times w \times 3}$, we compute the set $I_T = \{t(I) \mid t \in T \}$. We feed this set (sequentially or as a batch) into our model  $G(\boldsymbol{x}; \boldsymbol{\theta})$, which takes input $\boldsymbol{x}$ and has (frozen) weights $\boldsymbol{\theta}$. It has a patch size $P$ and hidden dimension $D$ so produces a set of features $F_{T} = \{G(i) \mid i \in I_T \}$. 

Each $f \in F_{T}$ has dimensions $f \in \mathbb{R}^{(h/P) \times (w/P) \times D}$ and so we nearest-neighbour resize each $f$ back to the original image dimensions to get $F'_{T}$ with elements $f' \in \mathbb{R}^{h \times w \times D}$. Next we apply each inverse transformation in $T^{-1} = \{t_{1}^{-1} , t_{2}^{-1} , ...\}$ to $F'_{T}$ to get $F'$. Finally we average over the $t$ in $F'_{T}$ to get $F \in \mathbb{R}^{h \times w \times D}$, our upsampled features for $I$.

As a concrete example, consider $T = \{t_{1}, t_{2}, ...\}$ to be the set of pixel-shift operators for each direction in a Moore neighbourhood for distances $\{d \mid 0 < d < P/2; d \in \mathbb{Z}\}$. When we compute the features of these shifted inputs, some of the spatial information from pixel $p$ will spill into neighbouring patches, such that when the features are resized and the inverse transformation (\textit{i.e,} the shift in the opposite direction) applied, the information is `put back in the right place' at pixel $p$.

This method extends to other transformations, including rotations and flips (useful for averaging away positional information) and arbitrary combinations thereof; this is realised via partial functions. We also extract high-resolution attention maps from the model at the same time in a similar manner. Similar to FeatUp, this approach upsamples the features by gaining new information from querying the model with transformations of the input - in FeatUp this information is encoded in a network (learned upsampling kernel or implict), in ours it is in a single forward pass.

There are several advantages: this method is model agnostic (assuming they use spatialised features), requires no additional training and works with the strided approach discussed in Section \ref{sec:feat_upsample}. The memory and time cost is linear with the number of transforms $T$ (see Section \ref{sec:supp_transforms_memory_cost}) if batched; for sequential processing the memory cost is constant (relative to the memory and time costs of the strided method).
Examples of the feature resolution can be seen in Sections \ref{sec:qual_compare} and \ref{sup:more_qual_compare}. Limitations are discussed in Section \ref{sec:limits}. Discussion on the impact of these transformations, as well as the number needed for a given stride is available in Section \ref{sec:supp_transforms}.

\textcolor{black}{In terms of the Vision Transformer whose features we upsample, we choose either DINO-S-8 or DINOv2-S-14. DINO-S-8 has a smaller patch size, and therefore blurs less when upsampled, but was trained on a smaller dataset and has less strong semantic distinctions than DINOv2. Other ViTs were tested (see Supplementary Figure \ref{fig:supp_model_compare}) but were fixed in their input size. We used the DINOv2 checkpoint with added register tokens, which prevents some background tokens from having anomalously high attention values \cite{REG_TOKENS}.}

\subsection{Unsupervised segmentation workflow}
\label{sec:unsupervised_method}
These high-resolution features can then be used directly (\textit{i.e,} without including additional colour information or self-training networks) in unsupervised downstream tasks. Our approach is to produce a `Class-Agnostic Segmentation' (CAS)\cite{CAS} - drawing parallels to class-agnostic object detection \cite{VIT_UNSUPERVISED_OD_SURVEY} - using only the information available in the features, and from that perform tasks like object localization, saliency detection and semantic segmentation.
A full diagram of the workflow is shown in Figure \ref{fig:explanation}.

We begin by clustering the high-resolution features into $C=80$ clusters using $k$-means clustering. 
To generate a foreground/background distinction we measure the `attention density', $\rho_{A}$, which is the attention per unit area of the \texttt{[CLS]} token in a cluster.
Clusters with $\rho_{A} > \Bar{\rho}_{A}$ are deemed foreground clusters. 
The use of $\rho_{A}$ is to create a well-separated distribution over the clusters, where background clusters are large and have low attention, so have a small resulting $\rho_{A}$. Not relying on the `largest class/cluster = background'\cite{LOST} or `most border pixels = background'\cite{MAE_VS_DINO_SSL} heuristics allows more flexibility on what is counted as foreground.  
This is especially useful when moving away from natural images towards less centralised, more homogeneous micrographs.

We then measure the cosine distance between each foreground and background cluster taking the modal distance as the `semantic distance', a measure of inter-class distance in feature space for the image $I$. That this semantic distance merging will produce a good segmentation assumes that the distance between a class in the foreground and a different one in the background is similar enough to the distance between two different classes in the foreground (\textit{i.e,} a bike vs. the woods and the rider vs. the bike).

Agglomerative merging with complete linkage is then used to merge clusters that have a distance less than the semantic distance, producing a rough CAS map of the image. 
This can be refined with a CRF\cite{EFFICIENT_CRF} to ameliorate the blurring problem. Multiplying the semantic distance by a scalar factor $\lambda$ allows control of the granularity of the merging - varied between 0.95 \textcolor{black}{(less merging)} and 1.1 \textcolor{black}{(more merging)} during experiments. \textcolor{black}{Increasing the number of initial clusters increases the time and memory complexity of the merging step, decreasing reduces the object-level resolution and makes the statistics of the cluster `attention density' less robust; we choose 80 as a middle ground.}

Combining the CAS map and attention information enables various downstream tasks: like object detection by drawing bounding boxes around unconnected regions in each class or  saliency detection via attention density of the CAS map. \textcolor{black}{For each unsupervised segmentation experiment in Sections \ref{sec:obj_localize_results} and \ref{sec:foreground_seg_results} we used DINOv2-S-14 with stride set to 4, shifts of 1 and 2px and flip transforms. The merge parameter $\lambda$ was set to 1.0. Stride 4 was the highest resolution achievable accounting for memory and time scaling, shifts greater than 2 had little affect on the feature resolution (see Section \ref{sec:supp_transforms}) and flip transformations averaged out some positional bias (see Figure \ref{fig:positional_bias}).}

Like the feature upsampling, this approach works per-image, for any model, and without extra training. 
We note that whilst FeatUp could be used for upsampling the features, it does not upsample the model attention maps at the same time, which are needed for the foreground/background separation.

\subsection{Weakly supervised segmentation}
\label{sec:method_wss}
Following existing work in the materials imaging community which train a random forest classifier\cite{WEKA, ILASTIK, NAPARI_APOC, NAPARI_FEATURE_CLASSIFIER} to map from classical image features to user labels, we train a classifier to map from our \textcolor{black}{upsampled pixel-wise} features from ViT models to user labels. These user labels are from `paint-brush' style annotation, (see Figure \ref{fig:wss} for an example) so may be sparse. Each $D$-dimensional vector that describes a (labelled) pixel is used alongside its associated label as a training example for the classifier. 

\textcolor{black}{We compare across three `feature-sets': Weka style classical features (Gaussian blurs, Sobel filter, Hessian filter, membrane projections, difference of Gaussians), upsampled DINOv2-S-14 with stride 4 and a `hybrid' scheme, where we concatenate the classical and upsampled ViT features in the channel dimension. This `hybrid' approach allows the classifier to leverage both the high-resolution (but inexpressive) classical features and the coarser, but more semantic upsampled ViT features. This hybrid approach is therefore a superset of the classical features, so should have strictly improved performance (assuming the upsampled ViT features add useful information).}

\textcolor{black}{Across the experiments in Section \ref{sec:wss} we focus on two types of classifier: random forests and logistic regression. We use a random forest classifier for the classical features to ensure similarity to existing schemes \cite{WEKA, ILASTIK, NAPARI_APOC, NAPARI_FEATURE_CLASSIFIER}. For the upsampled ViT features we also experiment with using logistic regression, chosen for its simplicity relative to random forests, with the goal of demonstrating that the more expressive (and non-linear) features required a less complex classifier. We justify this choice in Figure \ref{fig:supp_classifier_ablation}, which shows the segmentation performance when using upsampled ViT features to be fairly consistent across classifiers, and for the classical features to perform best when using random forests.}

\textcolor{black}{Unless otherwise stated, the random forest is `Weka-style', with 200 trees,  2 features per split and depth of 10. The logistic regression used default scikit-learn \cite{SKLEARN} parameters, with L2 regularization and 1000 iterations. Both classifiers had class-frequency weights applied. For ViT upsampling, we used DINOv2-S-14 with stride 4, shifts of 1 and 2px and flip transforms. }

\subsection{Datasets}

\textcolor{black}{We test our unsupervised segmentation workflow on two tasks: object localization - determining a bounding box for every foreground object in an image - and foreground object segmentation, which is predicting all pixels that belong to the foreground object(s) (\textit{i.e,} a binary segmentation). For the object localization task (Section \ref{sec:obj_localize_results}) we use the Video Object Classification (VOC) datasets, specifically VOC07 and VOC12\cite{VOC07, VOC12}. These are standard natural images of a range of everyday objects, where each image contains one or more examples of an object and the corresponding bounding boxes. It is also possible to use VOC as a semantic segmentation dataset, which we do to evaluate the performance of the upsampled features with linear probes in Section \ref{sec:linear_probes}}.

\textcolor{black}{For the foreground object segmentation (Section \ref{sec:foreground_seg_results}) we use two datasets: the Caltech-UCSD Birds dataset (CUBS)\cite{CUBS} and the Dalian University of Technology (DUTS) dataset\cite{DUTS}. CUBS contains images of various birds in difference contexts, alongside binary segmentations of the bird vs the background. DUTS contains images of various everyday objects and their segmentation. }

\textcolor{black}{When evaluating the performance in weakly-supervised segmentation in Section \ref{sec:wss} we use a hand-labelled dataset of Transmission Electron Micrographs (TEM) of human T-cells \cite{CELL_TEM_DATASET} with three classes: background, cell, nucleus. We also examine a series of micrographs from materials science taken with a variety of instruments. These images do not have a ground truth.}

\subsection{Evaluation Metrics}
\textcolor{black}{Segmentation performance was evaluated using Intersection over Union \textbf{(IoU)}, which for a binary segmentation (foreground vs background) measures the overlap of the predicted foreground pixels and ground truth label, divided by the total area of the two predictions. This ranges between 0 (no overlap) and 1 (predicted and ground truth are the same). For multiple classes, the IoU can be calculated separately for each class (by treating all other classes as background) and averaged, producing the mean Intersection over Union, or \textbf{mIoU}.}

\textcolor{black}{We measure object detection performance using `Correct Localization' or \textbf{CorLoc}, which is `true' if at least one predicted bounding boxes for an image has a greater than 50\% IoU with at least of the ground truth bounding boxes. This is averaged over all images in the dataset to get a score, which is the fraction of all images with a correct localization.}

\section{Results \& discussion}
\label{sec:results}

\subsection{Qualitative comparisons of features}
\label{sec:qual_compare}

\begin{figure}
\centering
    \includegraphics[width=1\linewidth]{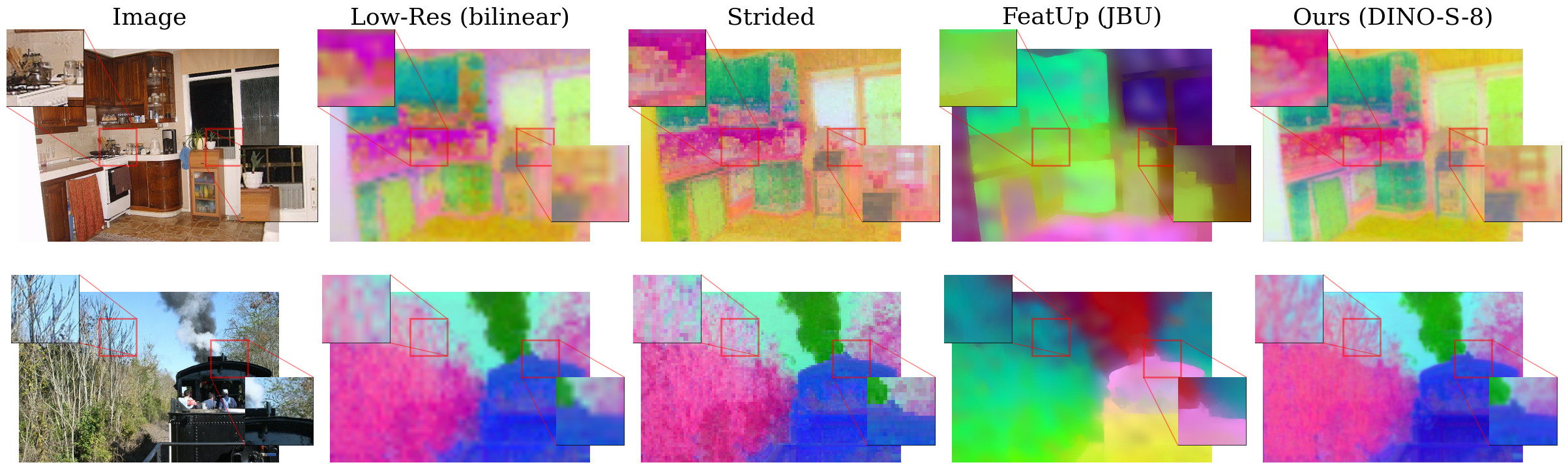}
    \caption{Feature resolution comparisons on two example images from VOC07 for different upsampling methods for DINO-S-8 features. FeatUp (JBU) is able to produce sharp edges for some objects but can blur others or miss fine details \textit{i.e,} the cups in the top image. Our method can capture such details, though the blurring introduced by using strided approach causes softer boundaries. Note the FeatUp featurizer was DINOv2-S-14.  }
    \label{fig:resolution_comparison}
\end{figure}

We compare the 3-component PCA of our upsampled features for DINO-S-8 qualitatively to the original low-resolution features, the result of setting the model stride to 4 and FeatUp in Figure \ref{fig:resolution_comparison}, noting an increase in resolution for fine-details and faithfulness to the original features. It should also be noted this approach works alongside the strided approach, and that the improvement from using our approach is relatively modest. 

Further comparisons are available in Figure \ref{fig:supp_method_compare}. Like FeatUp, our approach works across all models that produce spatialised features, this can be seen in Figure \ref{fig:supp_model_compare}. We discuss limitations, including blurring and boundary effects in Figure \ref{fig:supp_limits}. In each comparison we compare to FeatUp's fast, single-forward pass JBU approach as it is most similar to the aims of our approach - their implicit approach would produce higher-resolution features but would take far longer.

Following FeatUp we perform small-object retrieval\cite{FEATUP} in Figure \ref{fig:retrieval} - searching for the most similar point in a target image to a query point in another image where both images have been featurised with our upsampling approach.
A good match indicates the features are high-resolution and semantically relevant (\textit{i.e,} they are useful in downstream tasks). We add more keypoints to their image and track their matches, finding good agreement. When comparing to FeatUp (JBU), we find our feature similarities are better localised to the relevant objects (\textit{i.e.}, the traffic cones). A comparison using DINO-S-8 and showing the full target image is available in Figure \ref{fig:supp_retrieval}.   

\subsection{Quantitative feature comparisons via linear probes}
\label{sec:linear_probes}
\begin{table}
\centering
    \begin{tabular}{l c c }
        \toprule
        \textbf{Method} & \textbf{VOC12} \\
        \midrule
        Bilinear & 0.806 \\
        FeatUp (JBU) & \textbf{0.825} \\
        \midrule
        \textit{Ours (DINOv2-S-14)} & 0.701 \\
        \midrule
    \end{tabular}
    \caption{\textcolor{black}{mIoU on the VOC12 dataset when applying a linear classifier trained on bilinear-upsampled DINOv2-S-14 features to FeatUp upsampled features and features upsampled using our method (strided + shift transforms).}}
    \label{tab:sem_seg_probe}
\end{table}

To demonstrate the effectiveness of the learned features of DINOv2, a series of linear probes were trained to perform dense tasks like semantic segmentation and depth estimation given bilinear upsamplings of the patch features \cite{DINOv2}.
Assuming our method produces faithful and useful upsamplings, these linear probes should still be directly applicable to them and produce similar (if not better) results.
The results of these linear probes applied to our features for semantic segmentation on VOC2012\cite{VOC12} is reported in Table \ref{tab:sem_seg_probe}. 
There is a performance drop from applying the probe to the features produced by our method, this aligns with FeatUp\cite{FEATUP}'s findings for the strided method of higher resolution but noisier segmentation maps. 

\subsection{Unsupervised segmentation experiments}

\begin{center}
    \begin{figure}
    \centering
        \includegraphics[width=\linewidth]{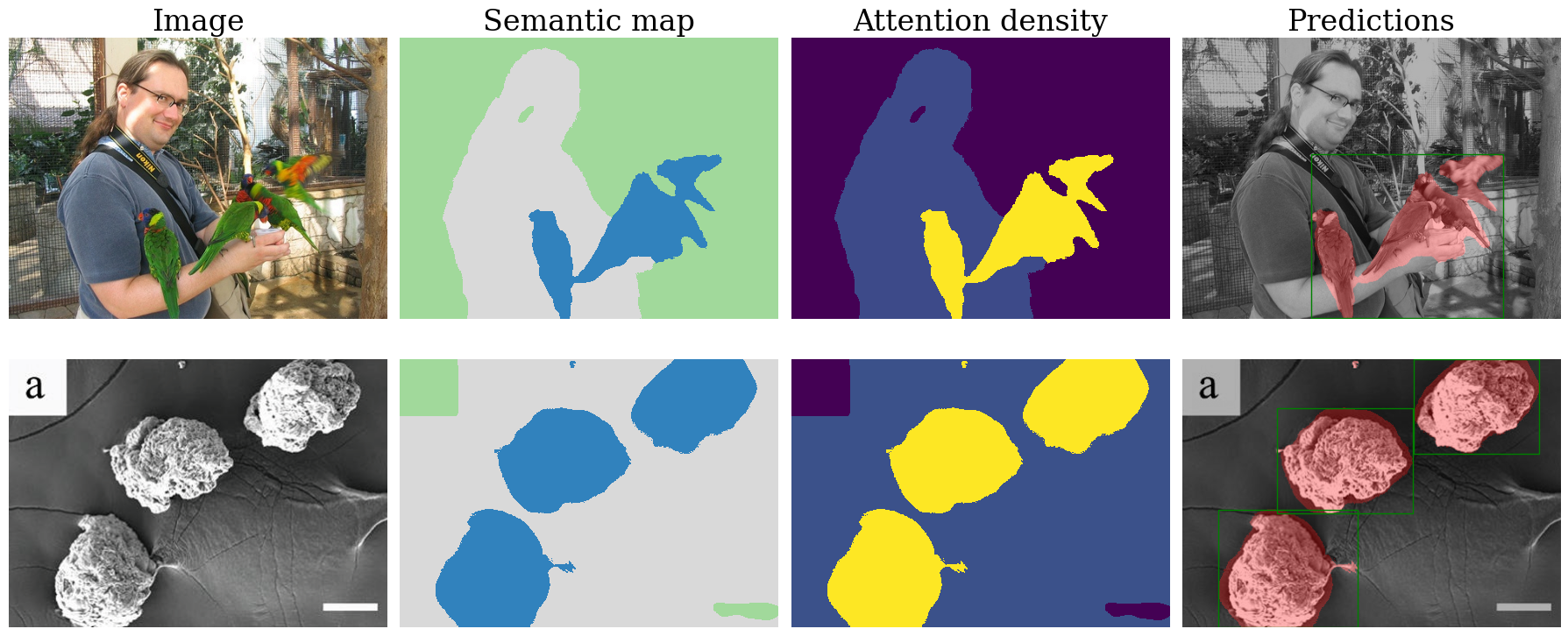}
        \caption{\textcolor{black}{Examples of the unsupervised segmentation workflow applied to a natural scale image and an SEM of mesoporous KIT-6 silica\cite{BLOBS}, showing how the unsupervised semantic map and attention densities can be used for object detection and localization.} } 
        \label{fig:unsupervised_exmaples}
    \end{figure}
\end{center}

\noindent\begin{minipage}[t]{0.5\textwidth}%
        \begin{tabular}[t]{l c c c }
            \toprule
             \textbf{Method} & \textbf{VOC07} & \textbf{VOC12} \\
            \midrule
            DINO \cite{DINO, VIT_UNSUPERVISED_OD_SURVEY} & 0.458 & 0.462  \\
            LOST \cite{LOST} & 0.620 & 0.640  \\
            Melas-Kyriazi \textit{et al.} \cite{DEEP_SPECTRAL} & 0.627 & 0.664  \\
            MOST \textit{et al.} (multi) \cite{MOST} & \textbf{0.748} & \textbf{0.774} \\ 
            \midrule
            \textit{Ours} \textcolor{black}{(DINOv2-S-14)} & 0.554  & 0.572 \\
            \textit{Ours} \textcolor{black}{(DINOv2-S-14, multi)}  & 0.718 & 0.725   \\
            \midrule
        \end{tabular}
        \captionof{table}{CorLoc on VOC07 \& VOC12 for various unsupervised object detection schemes. `Multi' refers to methods which produce multiple object bounding box predictions. Other values quoted from \cite{DEEP_SPECTRAL, MOST, VIT_UNSUPERVISED_OD_SURVEY} - we chose not to include methods which used a further self-training step.}
        \label{tab:obj_localize}
        
    \end{minipage}
    \hspace{0.05\linewidth}
    \begin{minipage}[t]{0.4\textwidth}
            \begin{tabular}[t]{l c c c }
                \toprule
                 \textbf{Method} & \textbf{CUB} & \textbf{DUTS} \\
                \midrule
                OneGAN \cite{ONEGAN}& 0.555 & -  \\
                Voynov \textit{et al.} \cite{VOYNOV} & 0.683 & 0.498  \\
                Melas-Kyriazi \textit{et al.} \cite{DEEP_SPECTRAL} & 0.769 & 0.514  \\
                MOST \cite{MOST} & - & 0.538 \\
                Deep Cut \cite{DEEPCUT} & 0.777 & 0.560 \\
                simSAM \cite{simSAM}& 0.770 & 0.582 \\
                TokenCut \cite{VIT_UNSUPERVISED_NORMCUT}& \textbf{0.795} & 0.624 \\
                \midrule
                \textit{Ours } \textcolor{black}{(DINOv2-S-14)} & 0.785 & \textbf{0.654}  \\
                \midrule
            
            \end{tabular}
            \captionof{table}{IoU of foreground object segmentation across the CUBS \& DUTS datasets. Other values quoted from \cite{DEEP_SPECTRAL, MOST, simSAM, VIT_UNSUPERVISED_OD_SURVEY}. \textcolor{black}{If a paper does not report a value for a dataset it has been replaced with a hyphen}.} 
            \label{tab:foreground_segment}
    \end{minipage}

\subsubsection{Object localization}
\label{sec:obj_localize_results}
For unsupervised object detection we simply take a bounding box around each connected component of each foreground class in the CAS map discussed in Section \ref{sec:unsupervised_method}. A class is a foreground class if its attention density is higher than the mean attention density - similar to the definition of foreground clusters. Our approach will occasionally decompose an object into multiple parts (\textit{i.e,} head/body, car/doors, \textit{etc.}) - to ameliorate this we introduce a `superbox' around the largest connected component of all foreground classes. If this superbox has more than an 80\% intersection over union (IoU) with another box, we retain only the other box. 

We then apply this approach to VOC07\cite{VOC07} and VOC12\cite{VOC12}, standard unsupervised object detection benchmarks, reporting the results in Table \ref{tab:obj_localize}.
The success metric is  `CorLoc' - the percentage of images where at least one of the predicted boxes has a greater than 50\% IoU with at least one of the ground truth boxes. We report two results: the CorLoc when only using the superbox (single object detection) and when using all predicted foreground boxes (multi-object detection).  Our method shows comparable performance to state-of-the-art in both single- and multi- \textcolor{black}{unsupervised} object detection.

\subsubsection{Foreground object segmentation}
\label{sec:foreground_seg_results}

For foreground object segmentation (also called `saliency detection') we use the same approach as in Sections \ref{sec:unsupervised_method} and \ref{sec:obj_localize_results}, producing a binary segmentation where all the foreground classes are counted as foreground, the rest as background. We apply our approach to two standard benchmark datasets, CUBS\cite{CUBS} and DUTS\cite{DUTS}, reporting the results in Table \ref{tab:foreground_segment}.
Our approach shows good performance on both datasets, surpassing state-of-the-art on DUTS. 

\subsection{Featuriser for zero-shot weakly supervised segmentation}
\label{sec:wss}
\begin{figure}[h]
\centering
    \includegraphics[width=1\linewidth]{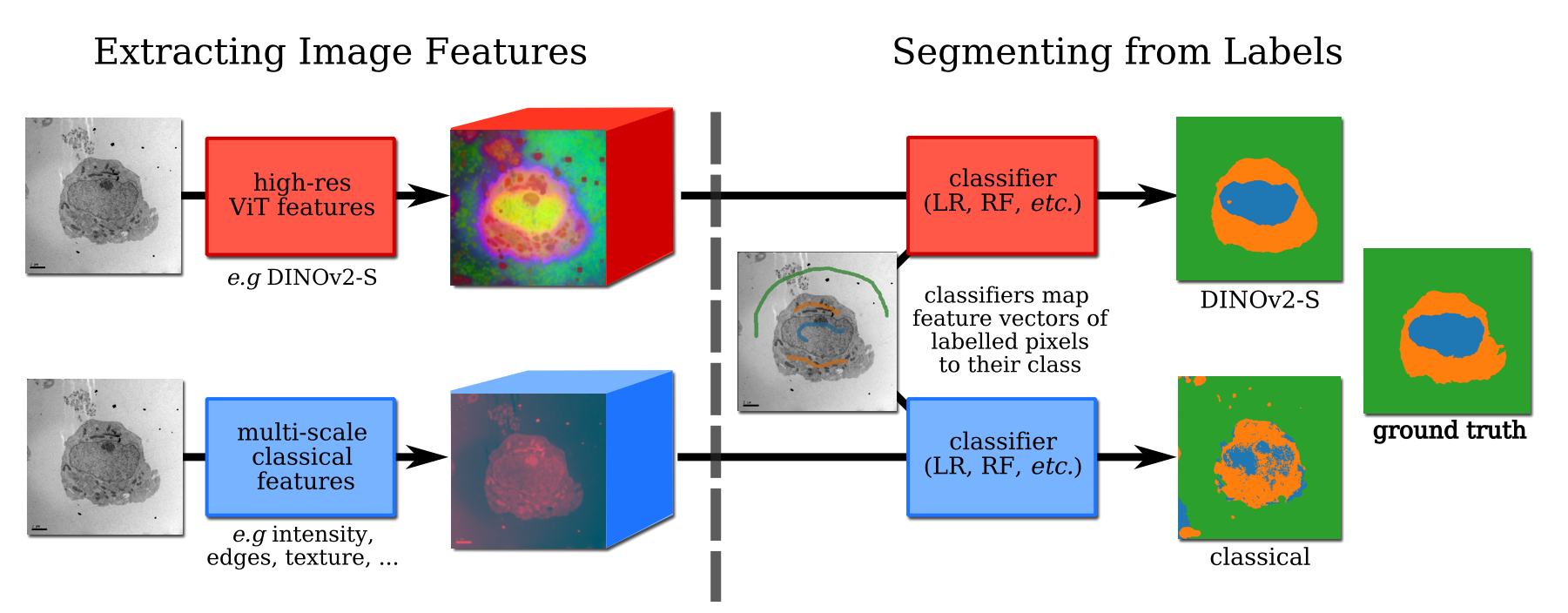}
    \caption{Explanation of the weakly supervised segmentation workflow, using an example of nucleic segmentation of a TEM image of a Jurkat human T-Cell\cite{CELL_TEM_DATASET}. Our method using high-resolution ViT features is able to capture the semantic information (interiority, foreground vs. background) needed for a good segmentation, compared to the classical image features + random forest approach. This disparity in feature richness can be seen in their respective feature spaces.}
    \label{fig:wss}
\end{figure}

\subsubsection*{Application to cell TEM dataset}
\label{sec:wss_t_cells}
Following Section \ref{sec:method_wss}, our aim is to use the richer semantics of these ViT models to perform complex semantic segmentation tasks; as an example we chose a dataset of 135 Transmission Electron Microscopy (TEM) micrographs of human T-cells\cite{CELL_TEM_DATASET} with three classes: background, cell, nucleus. We show an example in Figure \ref{fig:wss}, where the model using ViT features outperforms the classical method.

\begin{table}
\centering
    \begin{tabular}{l c c c}
        \toprule
         \textbf{Feature-set} & \textbf{mIoU} & \textbf{mIoU} (+CRF)\\
        \midrule
        Classical & 0.404 & 0.439 \\
        FeatUp (JBU) & 0.795 & 0.816 \\
        \textcolor{black}{DINOv2-S-14 (bilinear)} & \textcolor{black}{0.793} & \textcolor{black}{0.817} \\
        \midrule
        \textit{Ours} (DINO-S-8) & 0.776 & 0.803\\
        \textit{Ours }(DINOv2-S-14) & 0.797 & 0.827  \\
        \textit{Ours} (Hybrid) & 0.809 & \textbf{0.842} \\
        \midrule
    \end{tabular}
    \caption{mIoU of the three classes across the T-cell dataset for classifiers trained on the same set of labels with different pixel-features. Classifiers trained with upsampled ViT features (FeatUp, DINO-, hybrid) perform far better than when trained on classical features.}
    \label{tab:wss}
\end{table}

\begin{figure}[h]
\centering
    \includegraphics[width=1\linewidth]{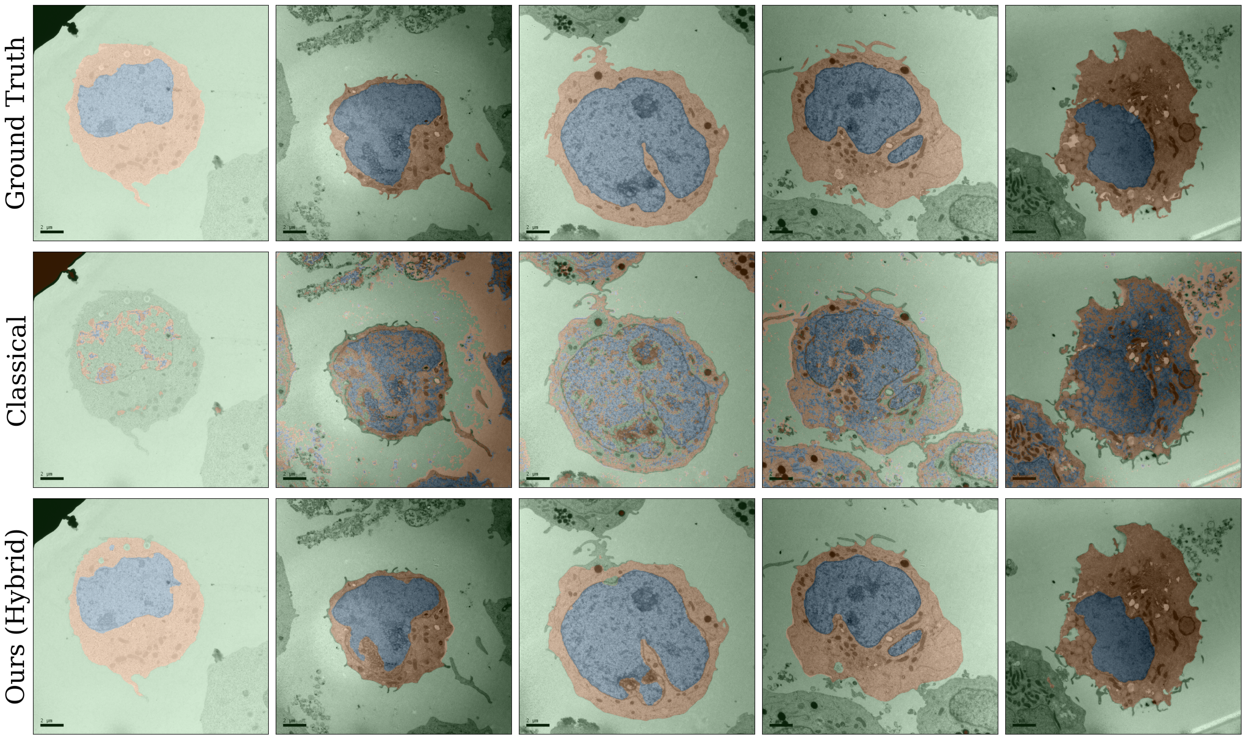}
    \caption{Predictions of the trained classifiers using classical or hybrid features on unlabelled examples from the T-cell dataset. The hybrid features produce good segmentations, demonstrating their ability to generalise well. Note a CRF was used for the hybrid cells and not for the classical.}
    \label{fig:wss_preds}
\end{figure}

Next we trained classifiers on partial labels across a set of six cells that cover the range of variation in the dataset, namely varying exposure, presence of background cells and multiple nuclei per cell. 
We then apply these trained classifiers to the rest of the (unlabelled) dataset and measure the mIoU for the three classes (background, nucleus, cell) relative to the ground truth annotations, presenting the results in Table \ref{tab:wss} and some example predictions on unseen data in Figure \ref{fig:wss_preds}. 
The stride of the ViT model was set to 4.
The labels and cells used to train the classifiers are available in Figure \ref{fig:supp_wss_labels}. 

We note a CRF\cite{EFFICIENT_CRF} was used to improve the segmentations for the case study, results without a CRF can be found in Section \ref{sec:supp_wss}.
Also included in Section \ref{sec:supp_wss} are more example segmentations on unlabelled cells, both with and without a CRF. Our method using the DINOv2 features shows markedly better performance across the dataset than the standard classical features + random forest approach. We attribute this to the richer feature space allowing concepts like interiority and foreground/background distinction to be expressed. A figure showing the resolution increase as the stride is reduced and transforms are added is available in Section \ref{sec:supp_transforms_wss}.

\subsubsection*{Results on diverse materials micrographs}

\begin{figure}
\centering
    \includegraphics[width=\linewidth]{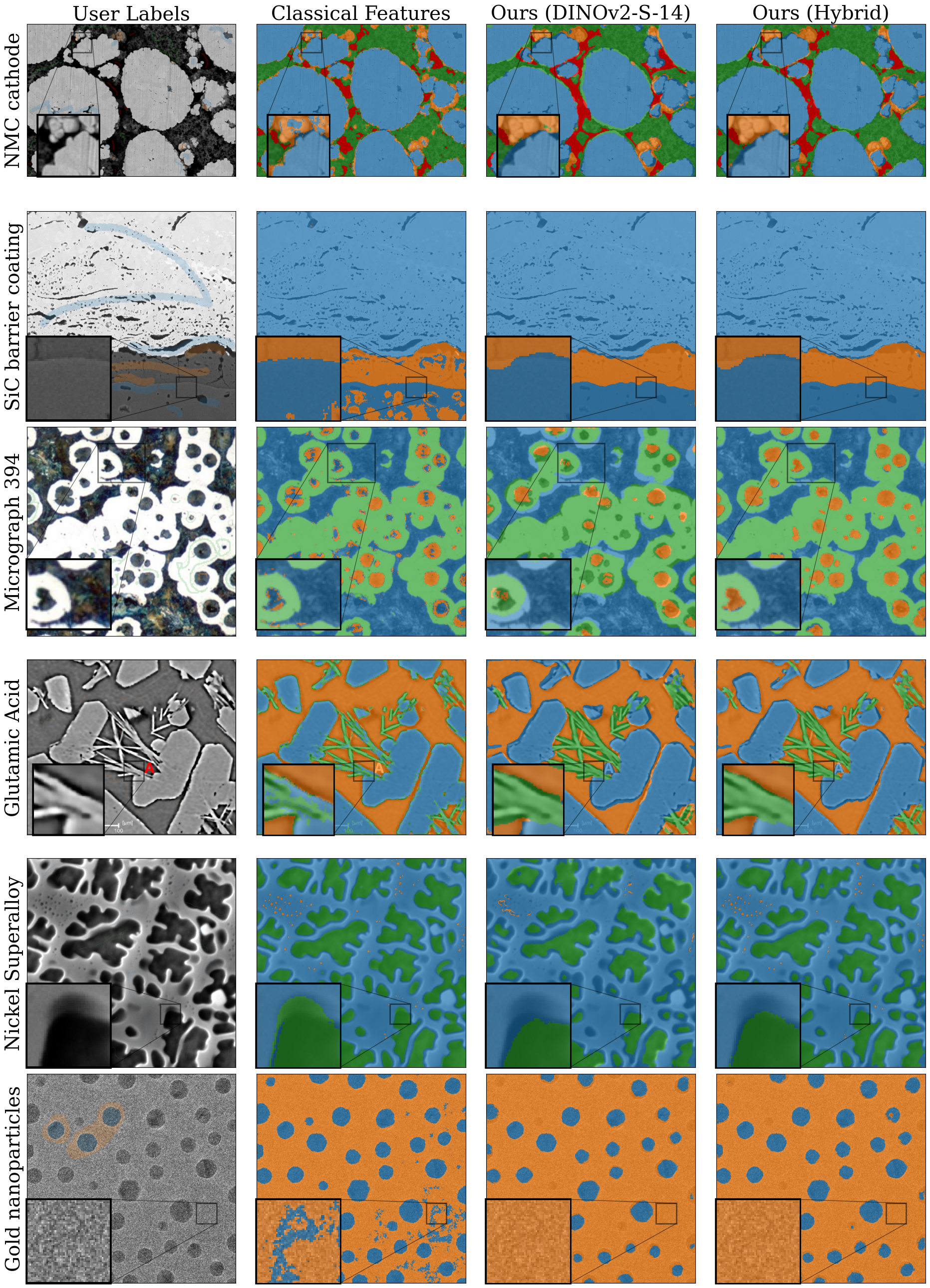}
    \caption{\textcolor{black}{Application of classical}, DINOv2 and classical + DINOv2 (hybrid) features for weakly supervised segmentation of materials such as NMC cathodes\cite{KINTSUGI, KINTSUGI_TFS}, SiC oxide layers\cite{SUPERALLOY}, cast iron alloys\cite{DOITPOMS_394}, glutamic acid polymorphs \cite{GLUTAMIC_ACID}, a nickel superalloy \cite{SUPERALLOY} and gold nanoparticles \cite{TEM_NANOPARTICLE}. The hybrid scheme is able to combine the strong semantic features of DINOv2 with the high-resolution classical features for good segmentations of complex tertiary and quaternary phases.}
    \label{fig:wss_examples}
    \vspace{-10pt}
\end{figure}

Despite the success of the CRF in correcting the (somewhat) blurry segmentations from the upsampled features for large phases like the cell or nucleus, we found it tended to remove small tertiary phases (\textit{i.e,} the organelles of the cell). This is a problem in materials science, where small, high frequency features like hairline cracks in battery materials are of great interest.   

To ameliorate this problem we experiment with concatenating the classical and DINOv2 features before training the classifier, in order to improve the accuracy on small, complex phases.  We call this the \say{hybrid} approach, and present the results in Figure \ref{fig:wss_examples} for six micrographs, again with the same set of labels and noting that a CRF was \textbf{not} used in these examples.

The first micrograph is from a high resolution Scanning Electron Microscope (SEM) image of lithium nickel manganese cobalt oxide cathode\cite{KINTSUGI, KINTSUGI_TFS} - the ViT features are able to distinguish between in-plane (light grey, flat) and out-of-plane (light grey, textured) active material (AM) present due to the `pore-back' effect\cite{KINTSUGI, KINTSUGI_TFS}. 
The ViT features also admit the distinction between voids inside an AM particle and the out-of-plane AM (which have similar textures).

The second micrograph is a silicon carbide environmental barrier coating \cite{SUPERALLOY}, where the goal is to segment the thermally-grown oxide layer in the middle. Similar to the cells example, the classical features struggle to separate the phases, which have similar greyscale intensity. The third micrograph is a Reflected Light Microscope (RFM) image of \say{cast iron with magnesium induced spheroidised graphite}\cite{DOITPOMS_394} - the hybrid features can distinguish between the interior graphite and occasionally similar-looking iron. 

The fourth micrograph is a slice from an X-ray Computed Tomography (XCT) of glutamic acid\cite{GLUTAMIC_ACID}, displaying two crystal polymorphs: the larger, blocky $\alpha$-polymorph and the smaller, needle-like $\beta$-polymorph\cite{GLUTAMIC_ACID}. The classical features are unable to distinguish between the two phases, struggling to delineate between edges of a large $\alpha$ block and the $\beta$ needles, whilst the DINOv2 features succeed but are blurred. The combination of the two produces a good segmentation of both phases - this ability to classify using structural information (rather than focussing on pixel-value) is useful in distinguishing between polymorphs for characterization techniques that have elemental contrast (like XCT or SEM), as the polymorphs will have similar pixel-values. 

The fifth micrograph is an SEM image of a nickel-based superalloy\cite{SUPERALLOY} with three phases: a large, connected matrix phase, secondary precipitates (large blobs) and tertiary precipitates (small blobs). Again, we find that the hybrid approach produces a better segmentation, especially of the third phase. The final micrograph is a high-resolution TEM image of gold nanoparticles \cite{TEM_NANOPARTICLE}, which displays noise and an exposure variation across the image that the classical features struggle to capture (given the localised labels).

We note that these experiments were performed using the default Weka feature set\cite{WEKA}; choosing a set more suited to the problem may improve performance, though this represents significant trial-and-error.
A random forest classifier was used for \textcolor{black}{all segmentations, and the labels were kept the same \textit{i.e,} the only difference between the columns is the feature-set used. } 
Labels were produced using the SAMBA web-app\cite{SAMBA}.
Discussion of the limitations of the deep features can be found in Section \ref{sec:limits}, focusing on the practical implications of using them in a materials context.

\section{Limitations}
\label{sec:limits}

\begin{figure}[h]
\centering
    \includegraphics[width=1\linewidth]{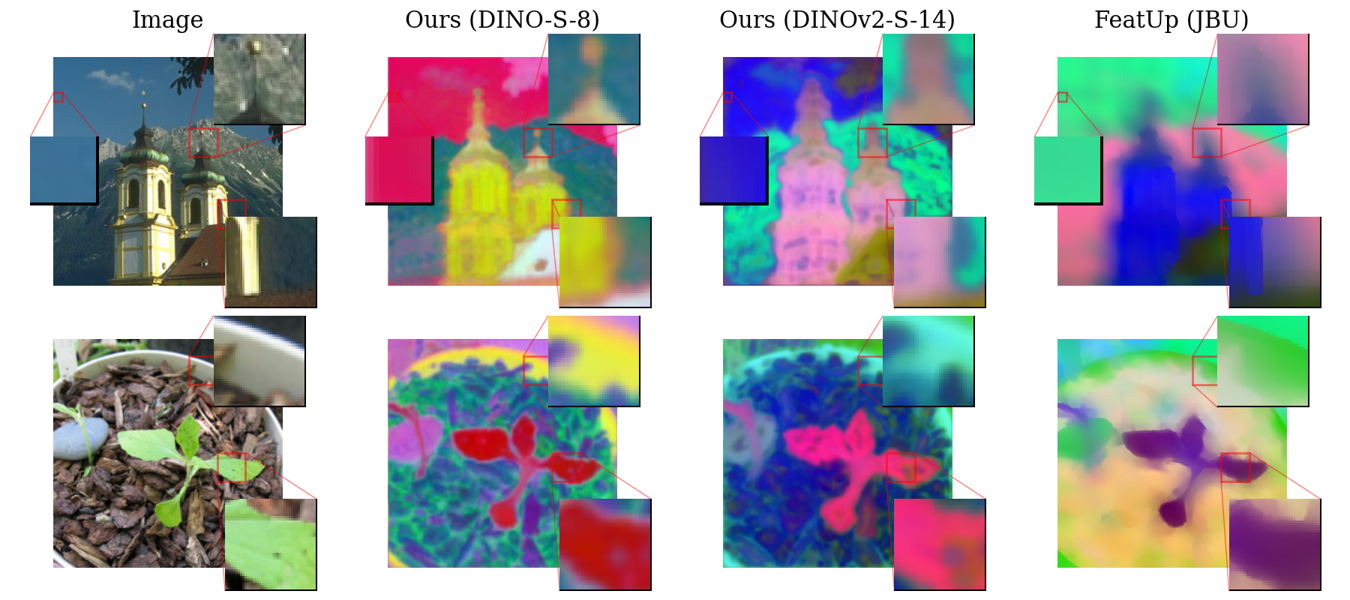}
    \caption{We highlight some of the limitations of our upsampling, primarily the blurring caused by reducing the model stride. This is reduced for models with a smaller base patch size (DINO-S-8 vs DINOv2-S-14), though a smaller base patch size incurs a higher memory cost; FeatUp (JBU) suffers some blurring, but retains sharp boundaries for foreground objects.}
    \label{fig:supp_limits}
\end{figure}

\begin{figure}[h]
\centering
    \includegraphics[width=0.8\linewidth]{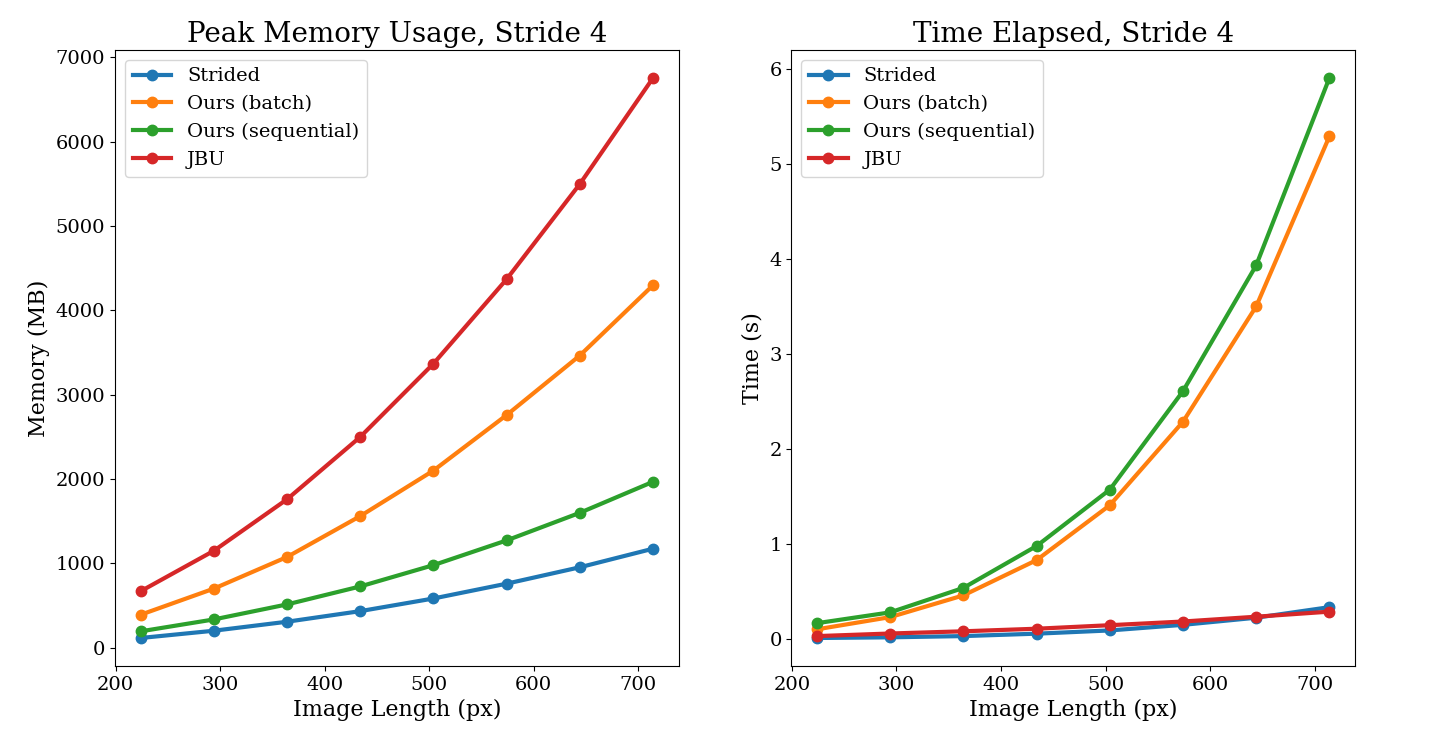}
    \caption{\textcolor{black}{Memory and time usage of various upsampling methods as function of square image length for DINOv2-S-14. We fix the stride for both strided and our methods to be 4 and use shift transforms with a distance of 2 and in an 8-neighbourhood for a total of 17 transformations. `Batch' refers to computing the features of the transformed image batch in one forward pass, `Sequential' to computing them one at a time. We note JBU has a harsher memory scaling as image length increases but far better time scaling. We achieve this better memory scaling than the expected $n^2$ via memory efficient attention\cite{MEM_EFF_ATTN}, which can be added post-hoc into any existing ViT network and admits $\sqrt{n}$ memory cost with $n$ tokens (though still an $n^2$ time cost).
    If the stride were to be increased beyond 4, the memory cost may exceed JBU, though this is undesirable due to numerical errors encountered at low strides/high upsampling factors. Float16 precision was used for both measurements; note for image length $l$, $n \propto l^2$. Values measured on an NVIDIA RTX A6000.}}
    \label{fig:supp_practical}
\end{figure}

The primary limitation of our feature upsampling method is the blurring introduced by the striding (see Figure \ref{fig:supp_limits}), which limits the resolution of the later downstream tasks. There are also practical problems with decreasing the model stride (and therefore increasing the number of tokens, $n$): although the memory cost does not scale $n^2$ thanks to memory-efficient attention\cite{MEM_EFF_ATTN}, the time-cost still scales with $n^2$ (\textit{i.e,} $N_\text{pixels}^4$).  

\textcolor{black}{We present a comparison of time and memory usage of the various upsampling techniques as function of image side length in Figure \ref{fig:supp_practical}. Our approach has a better memory scaling than JBU, but a worse time scaling; in an application context where memory is fixed, a higher time cost is preferable.} However, materials science frequently handles high resolution volumetric data, so reducing this featurisation time is important to ensure these data can be processed.

The unsupervised segmentation workflow tends to decompose large foreground objects into parts (head/body, \textit{etc.}) and treat those as separate classes - see Figure \ref{fig:supp_unsupervised_problems}. This could be improved by merging the features of individual pixels (rather than clusters), but this would be expensive computationally.

\begin{figure}
\centering
    \includegraphics[width=1\linewidth]{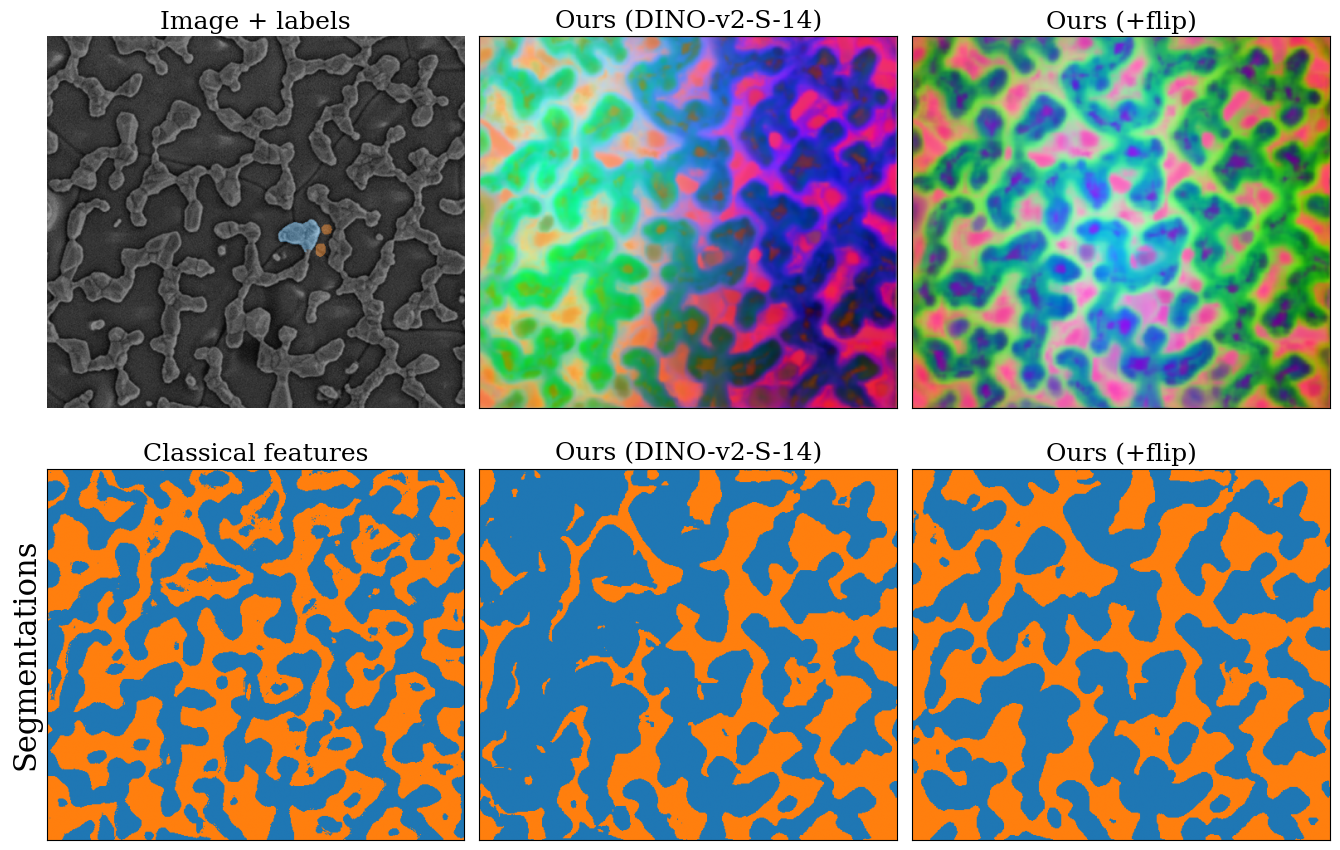}
    \caption{Demonstrating the positional biases present in DINOv2 features, and their consequences on weakly-supervised segmentation. The deep features, in contrast to the classical features, contain positional information, which the classifier can fit to (in the case of non-homogenous labels). Adding flip transformations can ameliorate this, though note there is still a center-image bias even after flip transforms have been applied. }
    \label{fig:positional_bias}
    \vspace{-10pt}
\end{figure}

The resolution problems and time-cost of generating the high-resolution features limit the applicability of the method in user-facing weakly supervised materials segmentation. Furthermore, it is possible for the classifiers to overfit to positional information present in the deep features - this requires additional homogeneous labelling to overcome, or some augmentations to reduce their importance, shown in Figure \ref{fig:positional_bias}. Empirically, combining classical and deep features can make the trained classifiers `brittle', \textit{i.e,} less sensitive to semantic information and therefore worse at generalizing.

\section{Conclusion}
\label{sec:conclusion}
To conclude, we have demonstrated a novel single-forward pass upsampling technique for features of vision transformers, like DINOv2. This was then applied to perform unsupervised object detection and segmentation, achieving strong results compared to baselines.
One potential use for this unsupervised workflow is generating high-quality semantic region proposals for interactive segmentation datasets \cite{SAM, CLUSTERING_FEATURES_FOR_DATASET}. 

Finally, we demonstrated the use of these upsampled features in weakly supervised materials segmentation, where they capture complex relationships that the current classical approaches are unable to express and achieve highly accurate segmentations. We expect that the ability to perform well across different materials, instruments and imaging conditions mean that the use of deep ViT features will greatly improve and expedite automated materials characterization.

There are many avenues for future work: improving the resolution and speed of feature extraction, utilizing the features in novel setups (\textit{i.e,} defect detection/classification) or using them for spatialised property prediction (regression) tasks. 

\section*{Acknowledgements}
This work was supported by funding from the the EPRSC and SFI Centre for Doctoral Training in Advanced Characterisation of Materials (EP/S023259/1 received by RD) and the Royal Society (IF\textbackslash R2\textbackslash 222059 received by AV as a Royal Society Industry Fellow).

\section*{Code and Data Availability}
\label{sec:data_availability}
The code needed to reproduce the results of the paper is available at \url{https://github.com/tldr-group/HR-Dv2}  with an MIT license agreement. The data for the case studies is available at \url{https://zenodo.org/records/14722134}.

\section*{References}
\addcontentsline{toc}{section}{References}

\def\addvspace#1{}

	\renewcommand{\refname}{ \vspace{-\baselineskip}\vspace{-1.1mm} }
	\bibliographystyle{ieeetr}
    \bibliography{main}

\begin{thebibliography}{10}

\bibitem{ATTN_IS_ALL_YOU_NEED}
A.~Vaswani, N.~Shazeer, N.~Parmar, J.~Uszkoreit, L.~Jones, A.~N. Gomez, L.~Kaiser, and I.~Polosukhin, ``Attention is all you need,'' {\em Advances in neural information processing systems}, vol.~30, 2017.

\bibitem{FM_SURVEY}
C.~Zhou, Q.~Li, C.~Li, J.~Yu, Y.~Liu, G.~Wang, K.~Zhang, C.~Ji, Q.~Yan, L.~He, H.~Peng, J.~Li, J.~Wu, Z.~Liu, P.~Xie, C.~Xiong, J.~Pei, P.~S. Yu, and L.~Sun, ``A {Comprehensive} {Survey} on {Pretrained} {Foundation} {Models}: {A} {History} from {BERT} to {ChatGPT},'' {\em arXiv preprint arXiv:2302.09419}, 2023.

\bibitem{BERT}
J.~Devlin, M.-W. Chang, K.~Lee, and K.~Toutanova, ``Bert: Pre-training of deep bidirectional transformers for language understanding,'' {\em arXiv preprint arXiv:1810.04805}, 2019.

\bibitem{T5}
C.~Raffel, N.~Shazeer, A.~Roberts, K.~Lee, S.~Narang, M.~Matena, Y.~Zhou, W.~Li, and P.~J. Liu, ``Exploring the limits of transfer learning with a unified text-to-text transformer,'' {\em arXiv preprint arXiv:1910.10683}, 2023.

\bibitem{GPT_1}
A.~Radford, K.~Narasimhan, T.~Salimans, and I.~Sutskever, ``Improving language understanding by generative pre-training,'' 2018.

\bibitem{GPT_3}
T.~B. Brown, B.~Mann, N.~Ryder, M.~Subbiah, J.~Kaplan, P.~Dhariwal, A.~Neelakantan, P.~Shyam, G.~Sastry, A.~Askell, S.~Agarwal, A.~Herbert-Voss, G.~Krueger, T.~Henighan, R.~Child, A.~Ramesh, D.~M. Ziegler, J.~Wu, C.~Winter, C.~Hesse, M.~Chen, E.~Sigler, M.~Litwin, S.~Gray, B.~Chess, J.~Clark, C.~Berner, S.~McCandlish, A.~Radford, I.~Sutskever, and D.~Amodei, ``Language {Models} are {Few}-{Shot} {Learners},'' {\em arXiv preprint arXiv:2005.14165}, 2020.

\bibitem{CLIP}
A.~Radford, J.~W. Kim, C.~Hallacy, A.~Ramesh, G.~Goh, S.~Agarwal, G.~Sastry, A.~Askell, P.~Mishkin, and J.~Clark, ``Learning transferable visual models from natural language supervision,'' in {\em International conference on machine learning}, pp.~8748--8763, PMLR, 2021.

\bibitem{DALLE}
A.~Ramesh, M.~Pavlov, G.~Goh, S.~Gray, C.~Voss, A.~Radford, M.~Chen, and I.~Sutskever, ``Zero-{Shot} {Text}-to-{Image} {Generation},'' {\em arXiv preprint arXiv:2102.12092}, 2021.

\bibitem{STABLE_DIFFUSION}
R.~Rombach, A.~Blattmann, D.~Lorenz, P.~Esser, and B.~Ommer, ``High-{Resolution} {Image} {Synthesis} with {Latent} {Diffusion} {Models},'' {\em arXiv preprint arXiv:2112.10752}, 2022.

\bibitem{SAM}
A.~Kirillov, E.~Mintun, N.~Ravi, H.~Mao, C.~Rolland, L.~Gustafson, T.~Xiao, S.~Whitehead, A.~C. Berg, W.-Y. Lo, P.~Dollár, and R.~Girshick, ``Segment {Anything},'' {\em arXiv preprint arXiv:2304.02643}, 2023.

\bibitem{YOLO}
J.~Redmon, S.~Divvala, R.~Girshick, and A.~Farhadi, ``You {Only} {Look} {Once}: {Unified}, {Real}-{Time} {Object} {Detection},'' {\em arXiv preprint arXiv:1506.02640}, 2016.

\bibitem{YOLO_REVIEW}
J.~Terven, D.-M. Córdova-Esparza, and J.-A. Romero-González, ``A {Comprehensive} {Review} of {YOLO} {Architectures} in {Computer} {Vision}: {From} {YOLOv1} to {YOLOv8} and {YOLO}-{NAS},'' {\em Machine Learning and Knowledge Extraction}, vol.~5, pp.~1680--1716, Nov. 2023.
\newblock Publisher: MDPI AG.

\bibitem{DINOv2}
M.~Oquab, T.~Darcet, T.~Moutakanni, H.~Vo, M.~Szafraniec, V.~Khalidov, P.~Fernandez, D.~Haziza, F.~Massa, A.~El-Nouby, M.~Assran, N.~Ballas, W.~Galuba, R.~Howes, P.-Y. Huang, S.-W. Li, I.~Misra, M.~Rabbat, V.~Sharma, G.~Synnaeve, H.~Xu, H.~Jegou, J.~Mairal, P.~Labatut, A.~Joulin, and P.~Bojanowski, ``{DINOv2}: {Learning} {Robust} {Visual} {Features} without {Supervision},'' {\em arXiv preprint arXiv: 2304.07193}, 2023.

\bibitem{DINO}
M.~Caron, H.~Touvron, I.~Misra, H.~Jégou, J.~Mairal, P.~Bojanowski, and A.~Joulin, ``Emerging {Properties} in {Self}-{Supervised} {Vision} {Transformers},'' 2021.

\bibitem{IJEPA}
M.~Assran, Q.~Duval, I.~Misra, P.~Bojanowski, P.~Vincent, M.~Rabbat, Y.~LeCun, and N.~Ballas, ``Self-{Supervised} {Learning} from {Images} with a {Joint}-{Embedding} {Predictive} {Architecture},'' {\em arXiv preprint arXiv: 2301.08243}, 2023.

\bibitem{DENSE_VIT_FEATURES}
S.~Amir, Y.~Gandelsman, S.~Bagon, and T.~Dekel, ``Deep {ViT} {Features} as {Dense} {Visual} {Descriptors},'' 2022.

\bibitem{VIT_UNSUPERVISED_OD_SURVEY}
O.~Siméoni, E.~Zablocki, S.~Gidaris, G.~Puy, and P.~Pérez, ``Unsupervised {Object} {Localization} in the {Era} of {Self}-{Supervised} {ViTs}: {A} {Survey},'' {\em arXiv preprint arXiv: 2310.12904}, 2023.

\bibitem{MOST}
S.~S. Rambhatla, I.~Misra, R.~Chellappa, and A.~Shrivastava, ``{MOST}: {Multiple} {Object} localization with {Self}-supervised {Transformers} for object discovery,'' {\em arXiv preprint arXiv: 2304.05387}, 2023.

\bibitem{LOST}
O.~Siméoni, G.~Puy, H.~V. Vo, S.~Roburin, S.~Gidaris, A.~Bursuc, P.~Pérez, R.~Marlet, and J.~Ponce, ``Localizing {Objects} with {Self}-{Supervised} {Transformers} and no {Labels},'' {\em arXiv preprint arXiv: 2109.14279}, 2021.

\bibitem{VIT_UNSUPERVISED_NORMCUT}
Y.~Wang, X.~Shen, S.~Hu, Y.~Yuan, J.~Crowley, and D.~Vaufreydaz, ``Self-{Supervised} {Transformers} for {Unsupervised} {Object} {Discovery} using {Normalized} {Cut},'' {\em arXiv preprint arXiv: 2202.11539}, 2022.

\bibitem{DEEP_SPECTRAL}
L.~Melas-Kyriazi, C.~Rupprecht, I.~Laina, and A.~Vedaldi, ``Deep {Spectral} {Methods}: {A} {Surprisingly} {Strong} {Baseline} for {Unsupervised} {Semantic} {Segmentation} and {Localization},'' {\em arXiv preprint arXiv: 2205.07839}, 2022.

\bibitem{MATERIALS_DATA_DIVERSITY}
A.~Goetz, A.~R. Durmaz, M.~Müller, A.~Thomas, D.~Britz, P.~Kerfriden, and C.~Eberl, ``Addressing materials’ microstructure diversity using transfer learning,'' {\em npj Computational Materials}, vol.~8, p.~27, Feb. 2022.

\bibitem{WEKA}
I.~Arganda-Carreras, V.~Kaynig, C.~Rueden, K.~W. Eliceiri, J.~Schindelin, A.~Cardona, and H.~Sebastian~Seung, ``Trainable {Weka} {Segmentation}: a machine learning tool for microscopy pixel classification,'' {\em Bioinformatics}, vol.~33, pp.~2424--2426, Aug. 2017.

\bibitem{ILASTIK}
S.~Berg, D.~Kutra, T.~Kroeger, C.~N. Straehle, B.~X. Kausler, C.~Haubold, M.~Schiegg, J.~Ales, T.~Beier, M.~Rudy, K.~Eren, J.~I. Cervantes, B.~Xu, F.~Beuttenmueller, A.~Wolny, C.~Zhang, U.~Koethe, F.~A. Hamprecht, and A.~Kreshuk, ``ilastik: interactive machine learning for (bio)image analysis,'' {\em Nature Methods}, Sept. 2019.

\bibitem{NAPARI_APOC}
R.~Haase, D.~Lee, D.~D. Pop, and L.~Žigutytė, ``{haesleinhuepf/napari-accelerated-pixel-and-object-classification: 0.14.1},'' Nov. 2023.

\bibitem{NAPARI_FEATURE_CLASSIFIER}
J.~Luethi and M.~Hess, ``{napari-feature-classifier: An interactive classifier plugin to use with label images and feature measurements}.''

\bibitem{FEATUP}
S.~Fu, M.~Hamilton, L.~Brandt, A.~Feldman, Z.~Zhang, and W.~T. Freeman, ``{FeatUp}: {A} {Model}-{Agnostic} {Framework} for {Features} at {Any} {Resolution},'' {\em arXiv preprint arXiv: 2403.10516}, 2024.

\bibitem{VIT}
A.~Dosovitskiy, L.~Beyer, A.~Kolesnikov, D.~Weissenborn, X.~Zhai, T.~Unterthiner, M.~Dehghani, M.~Minderer, G.~Heigold, S.~Gelly, J.~Uszkoreit, and N.~Houlsby, ``An {Image} is {Worth} 16x16 {Words}: {Transformers} for {Image} {Recognition} at {Scale},'' {\em arXiv preprint arXiv: 2010.11929}, 2021.

\bibitem{FOURIER_PE}
M.~Tancik, P.~P. Srinivasan, B.~Mildenhall, S.~Fridovich-Keil, N.~Raghavan, U.~Singhal, R.~Ramamoorthi, J.~T. Barron, and R.~Ng, ``Fourier {Features} {Let} {Networks} {Learn} {High} {Frequency} {Functions} in {Low} {Dimensional} {Domains},'' {\em arXiv preprint arXiv: 2006.10739}, 2020.

\bibitem{MAE}
K.~He, X.~Chen, S.~Xie, Y.~Li, P.~Dollár, and R.~Girshick, ``Masked {Autoencoders} {Are} {Scalable} {Vision} {Learners},'' {\em arXiv preprint arXiv: 2111.06377}, 2021.

\bibitem{AUDIO_MAE}
P.-Y. Huang, H.~Xu, J.~Li, A.~Baevski, M.~Auli, W.~Galuba, F.~Metze, and C.~Feichtenhofer, ``Masked autoencoders that listen,'' {\em arXiv preprint arXiv: 2207.06405}, 2023.

\bibitem{VIDEO_MAE}
Z.~Tong, Y.~Song, J.~Wang, and L.~Wang, ``Videomae: Masked autoencoders are data-efficient learners for self-supervised video pre-training,'' {\em arXiv preprint arXiv: 2203.12602}, 2022.

\bibitem{VJEPA}
A.~Bardes, Q.~Garrido, I.~Misra, J.~Ponce, X.~Chen, M.~Rabbat, Y.~LeCun, M.~Assran, and N.~Ballas, ``Revisiting feature prediction for learning visual representations from video,'' {\em preprint}, 2024.

\bibitem{DEIT}
H.~Touvron, M.~Cord, M.~Douze, F.~Massa, A.~Sablayrolles, and H.~Jégou, ``Training data-efficient image transformers \& distillation through attention,'' {\em arXiv preprint arXiv: 2012.12877}, 2021.

\bibitem{IBOT}
J.~Zhou, C.~Wei, H.~Wang, W.~Shen, C.~Xie, A.~Yuille, and T.~Kong, ``{iBOT}: {Image} {BERT} {Pre}-{Training} with {Online} {Tokenizer},'' {\em arXiv preprint arXiv: 2111.07832}, 2022.

\bibitem{VIT_ROBUST}
S.~Paul and P.-Y. Chen, ``Vision transformers are robust learners,'' {\em arXiv preprint arXiv:2105.07581}, 2021.

\bibitem{MAE_VS_DINO_SSL}
P.~Engstler, L.~Melas-Kyriazi, C.~Rupprecht, and I.~Laina, ``Understanding {Self}-{Supervised} {Features} for {Learning} {Unsupervised} {Instance} {Segmentation},'' {\em arXiv preprint arXiv: 2311.14665}, 2023.

\bibitem{JBU}
J.~Kopf, M.~F. Cohen, D.~Lischinski, and M.~Uyttendaele, ``Joint bilateral upsampling,'' {\em ACM Trans. Graph.}, vol.~26, p.~96–es, jul 2007.

\bibitem{CLUSTERING_UNSUPERVISED}
Y.~S.~J. Cheung, X.~Chen, L.~Yang, and H.~Zhao, ``A {Lightweight} {Clustering} {Framework} for {Unsupervised} {Semantic} {Segmentation},'' {\em arXiv preprint arXiv: 2311.18628}, 2023.

\bibitem{DEEPCUT}
A.~Aflalo, S.~Bagon, T.~Kashti, and Y.~Eldar, ``Deepcut: Unsupervised segmentation using graph neural networks clustering,'' {\em arXiv preprint arXiv: 2212.05853}, 2023.

\bibitem{TOKENCUT}
Y.~Wang, X.~Shen, Y.~Yuan, Y.~Du, M.~Li, S.~X. Hu, J.~L. Crowley, and D.~Vaufreydaz, ``Tokencut: Segmenting objects in images and videos with self-supervised transformer and normalized cut,'' {\em arXiv preprint arXiv: 2209.00383}, 2023.

\bibitem{SCRIBBLESEG}
X.~Chen, Y.~S.~J. Cheung, S.-N. Lim, and H.~Zhao, ``Scribbleseg: Scribble-based interactive image segmentation,'' {\em arXiv preprint arXiv: 2303.11320}, 2023.

\bibitem{WSS_GAN}
N.~Souly, C.~Spampinato, and M.~Shah, ``Semi and weakly supervised semantic segmentation using generative adversarial network,'' {\em arXiv preprint arXiv: 1703.09695}, 2017.

\bibitem{WSS_TRANSFORMER}
L.~Ru, Y.~Zhan, B.~Yu, and B.~Du, ``Learning affinity from attention: End-to-end weakly-supervised semantic segmentation with transformers,'' {\em arXiv preprint arXiv: 2203.02664}, 2022.

\bibitem{WSS_REVIEW}
L.~Chan, M.~S. Hosseini, and K.~N. Plataniotis, ``A comprehensive analysis of weakly-supervised semantic segmentation in different image domains,'' {\em International Journal of Computer Vision}, vol.~129, p.~361–384, Sept. 2020.

\bibitem{AL_REVIEW}
B.~Settles, ``Active {Learning} {Literature} {Survey},'' 2010.

\bibitem{REG_TOKENS}
T.~Darcet, M.~Oquab, J.~Mairal, and P.~Bojanowski, ``Vision transformers need registers,'' 2024.

\bibitem{CAS}
C.~Zhang, G.~Lin, F.~Liu, R.~Yao, and C.~Shen, ``Canet: Class-agnostic segmentation networks with iterative refinement and attentive few-shot learning,'' {\em arXiv preprint arXiv: 1903.02351}, 2019.

\bibitem{EFFICIENT_CRF}
P.~Krähenbühl and V.~Koltun, ``Efficient {Inference} in {Fully} {Connected} {CRFs} with {Gaussian} {Edge} {Potentials},'' {\em arXiv preprint arXiv: 1210.5644}, 2012.

\bibitem{SKLEARN}
F.~Pedregosa, G.~Varoquaux, A.~Gramfort, V.~Michel, B.~Thirion, O.~Grisel, M.~Blondel, P.~Prettenhofer, R.~Weiss, V.~Dubourg, {\em et~al.}, ``Scikit-learn: Machine learning in python,'' {\em Journal of machine learning research}, vol.~12, no.~Oct, pp.~2825--2830, 2011.

\bibitem{VOC07}
M.~Everingham, L.~Van~Gool, C.~K.~I. Williams, J.~Winn, and A.~Zisserman, ``The {PASCAL} {V}isual {O}bject {C}lasses {C}hallenge 2007 {(VOC2007)} {R}esults.'' http://www.pascal-network.org/challenges/VOC/voc2007/workshop/index.html.

\bibitem{VOC12}
M.~Everingham, L.~Van~Gool, C.~K.~I. Williams, J.~Winn, and A.~Zisserman, ``The {PASCAL} {V}isual {O}bject {C}lasses {C}hallenge 2012 {(VOC2012)} {R}esults.'' http://www.pascal-network.org/challenges/VOC/voc2012/workshop/index.html.

\bibitem{CUBS}
C.~Wah, S.~Branson, P.~Welinder, P.~Perona, and S.~Belongie, ``Caltech-ucsd birds-200-2011 (cub-200-2011),'' Tech. Rep. CNS-TR-2011-001, California Institute of Technology, 2011.

\bibitem{DUTS}
L.~Wang, H.~Lu, Y.~Wang, M.~Feng, D.~Wang, B.~Yin, and X.~Ruan, ``Learning to detect salient objects with image-level supervision,'' in {\em CVPR}, 2017.

\bibitem{CELL_TEM_DATASET}
V.~Morath, M.~Keuper, M.~Rodriguez-Franco, S.~Deswal, G.~Fiala, B.~Blumenthal, D.~Kaschek, J.~Timmer, G.~Neuhaus, S.~Ehl, O.~Ronneberger, and W.~W.~A. Schamel, ``Semi-automatic determination of cell surface areas used in systems biology.,'' {\em Frontiers in bioscience (Elite edition)}, vol.~5, pp.~533--545, Jan. 2013.
\newblock Place: Singapore.

\bibitem{BLOBS}
T.~Li, ``Particle size and pore size selection on ordered mesoporous silica,'' {\em ChemRxiv}, 2023.

\bibitem{ONEGAN}
Y.~Benny and L.~Wolf, {\em OneGAN: Simultaneous Unsupervised Learning of Conditional Image Generation, Foreground Segmentation, and Fine-Grained Clustering}, p.~514–530.
\newblock Springer International Publishing, 2020.

\bibitem{VOYNOV}
A.~Voynov, S.~Morozov, and A.~Babenko, ``Object segmentation without labels with large-scale generative models,'' {\em arXiv preprint arXiv: 2006.04988}, 2021.

\bibitem{simSAM}
C.~G. Kamra, I.~D. Mastan, N.~Kumar, and D.~Gupta, ``Simsam: Simple siamese representations based semantic affinity matrix for unsupervised image segmentation,'' {\em arXiv preprint arXiv: 2406.07986}, 2024.

\bibitem{KINTSUGI}
S.~J. Cooper, S.~A. Roberts, Z.~Liu, and B.~Winiarski, ``Methods—{Kintsugi} {Imaging} of {Battery} {Electrodes}: {Distinguishing} {Pores} from the {Carbon} {Binder} {Domain} using {Pt} {Deposition},'' {\em Journal of The Electrochemical Society}, vol.~169, p.~070512, July 2022.
\newblock Publisher: IOP Publishing.

\bibitem{KINTSUGI_TFS}
B.~Winiarski and P.~Barthelemy, ``Kintsugi imaging of battery electrodes with plasma {FIB}-{SEM}.''

\bibitem{SUPERALLOY}
J.~Stuckner, B.~Harder, and T.~M. Smith, ``Microstructure segmentation with deep learning encoders pre-trained on a large microscopy dataset,'' {\em npj Computational Materials}, vol.~8, p.~200, Sept. 2022.

\bibitem{DOITPOMS_394}
R.~F. Cochrane, ``Micrograph 394,'' 2002.
\newblock https://www.doitpoms.ac.uk/miclib/micrograph\_record.php?id=394.

\bibitem{GLUTAMIC_ACID}
T.~D. Turner, P.~Gajjar, I.~S. Fragkopoulos, J.~Carr, T.~T.~H. Nguyen, D.~Hooper, F.~Clarke, N.~Dawson, P.~J. Withers, and K.~J. Roberts, ``Measuring the {Particle} {Packing} of l-{Glutamic} {Acid} {Crystals} through {X}-ray {Computed} {Tomography} for {Understanding} {Powder} {Flow} and {Consolidation} {Behavior},'' {\em Crystal Growth \& Design}, vol.~20, no.~7, pp.~4252--4263, 2020.

\bibitem{TEM_NANOPARTICLE}
J.~P. Horwath, D.~N. Zakharov, R.~Mégret, and E.~A. Stach, ``Understanding important features of deep learning models for segmentation of high-resolution transmission electron microscopy images,'' {\em npj Computational Materials}, vol.~6, p.~108, July 2020.

\bibitem{SAMBA}
R.~Docherty, I.~Squires, A.~Vamvakeros, and S.~J. Cooper, ``{SAMBA}: {A} {Trainable} {Segmentation} {Web}-{App} with {Smart} {Labelling},'' {\em Journal of Open Source Software}, vol.~9, no.~98, p.~6159, 2024.
\newblock Publisher: The Open Journal.

\bibitem{MEM_EFF_ATTN}
M.~N. Rabe and C.~Staats, ``Self-attention does not need $o(n^2)$ memory,'' {\em arXiv preprint arXiv: 2112.05682}, 2022.

\bibitem{CLUSTERING_FEATURES_FOR_DATASET}
K.~Li, Y.~Zhao, Z.~Wang, Z.~Cheng, P.~Jin, X.~Ji, L.~Yuan, C.~Liu, and J.~Chen, ``Multi-granularity interaction simulation for unsupervised interactive segmentation,'' {\em arXiv preprint arXiv: {2303.13399}}, 2023.

\end{thebibliography}

\include{Sup_Info}
\newpage
\section*{Supplementary}
\setcounter{section}{0}
\setcounter{figure}{0}
\setcounter{table}{0}
\renewcommand*{\theHsection}{S.\the\value{section}}

\makeatletter
\renewcommand \thesection{S\@arabic\c@section}
\renewcommand\thetable{S\@arabic\c@table}
\renewcommand \thefigure{S\@arabic\c@figure}
\makeatother

\section{Full algorithms}

\begin{algorithm*}
\begin{algorithmic}[1]
\Require  Image $I$ (of size $(C,H,W)$), a ViT feature extraction model $\textbf{G}$ with hidden dimension $k$ and number of attention heads $N_{h}$, a set of image transformations (pixel shifts, flips) as partial functions $T$ with $N_T=|T|$, which last-layer attention vector (if any) to upsample $(\textbf{q}\text{uery}, \textbf{k}\text{ey}, \textbf{v}\text{alue or } \textbf{o}\text{utput})$. Note this is the attention map for the global $\texttt{[CLS]}$ token.
\Preprocessing Reduce stride of $\textbf{G}$'s input patch projection layer from 16 or 14 down to some $S$ (usually $S=4$).
\Statex
    \State $I_T$, shape $(N_T,C,H,W)$ $\leftarrow$ Apply all transformations in $T$ to $I$. 
    \State $F_T$, shape $(N_T,k + N_h,H/S,W/S)$ $\leftarrow$ Compute features of $I_T$ with $\textbf{G}$  \textit{i.e,} $\textbf{G}(i)\:\forall\:i \in I_T$ (attention is necessarily computed at the same time and stored inside $F_T$)
    \State $F'_T$, shape $(N_T,k + N_h,H,W)$  $\leftarrow$ Nearest-neighbor resize $F_T$ from ($H/S,W/S$) to ($H,W$)
    \State $F'$ $\leftarrow$ Apply the inverse of all transforms in $T$ to $F'_T$ \textit{i.e,} unshift and unflip
    \State $F$, shape $(k,H,W)$ \& $A$, shape $(N_h,H,W)$ $\leftarrow$ Average over all $f \in F'$ up to $k^{th}$ channel for features, and average all $f \in F'$ from $k^{th}$ to $k+N_h^{th}$ channel for attention.
    
    \Return $F$, the upsampled ViT features of image $I$, $A$ the upsampled attention maps of $I$. 
    \caption{{\footnotesize \textcolor{black}{Transformation based ViT feature upsampling (batched)} }}
    \label{alg:upsampling}
\end{algorithmic}
\end{algorithm*}

\begin{algorithm*}
\begin{algorithmic}[1]
\Require  Image $I$ (of size $(C,H,W)$), number of k-means clusters $C$, associated upsampled features $F$ (shape $(k,H,W)$) and attention maps $A$ (shape $(N_h,H,W)$) of the $\texttt{[CLS]}$ token. Optional number of classes $N_c$ and merge threshold multiplier $\lambda$.
\Preprocessing Sum the attention maps across the heads to produce $A'$, shape $(H,W)$
\Statex
    \State $\mu \leftarrow$ K-Means $F$ into $C$ clusters, optionally over $N_s$ samples
    \State $S^\mu$, shape (H,W) $\leftarrow$ assign each pixel in $F$ to its nearest cluster in $\mu$ to get an over-segmentation \textit{i.e,} $S^{\mu}_{i,j} \in [1, 2, ..., C]$
    \State $\rho_A$ $\leftarrow$ Compute attention density (sum of $A'$ where $S^{\mu} = i$ divided by number of pixels where $S^{\mu} = i$ ) for each cluster $i$ in $\mu$
    \State $\mu_{FG}$, $\mu_{BG}$ $\leftarrow$ Separate foreground (FG) and background (BG) clusters based on whether their attention density is larger than the mean attention density \textit{i.e,} $\rho_A > \bar{\rho_a}$
    \State $d_s$ $\leftarrow$ Find semantic distance $d_s$, the modal cosine distance between the features (in $F$) of every $\mu_{FG}$ and $\mu_{BG}$ multiplied by the threshold multiplier $\lambda$
    \If {$N_c$ supplied}
        \State $S_M$ $\leftarrow$ Greedily merge clusters in $\mu$ based on cosine distance until $N_c$ clusters remain
    \Else
        \State $S_M$ $\leftarrow$ Greedily merge clusters in $\mu$ if their cosine distance is less than the semantic distance \textit{i.e}, $\text{dist}(F_{\mu_i}, F_{\mu_j})$ < $d_s$ to get segmentation
    \EndIf
    \State $S_{M,CRF}$ $\leftarrow$ apply CRF based post-processing to refine $S_M$ based on cosine distance of a pixel's features from the cluster centroid relative to its colour similarity to neighbouring pixels   

    \Return $S_{M,CRF}$, a (H,W) semantic segmentation of image $I$ where the number of classes has been automatically determined based on its upsampled attention and feature maps. Note we can extract a foreground vs background segmentation from $\mu_{FG}$, $\mu_{BG}$.
    \caption{{\footnotesize \textcolor{black}{Unsupervised ViT feature-clustering based segmentation} }}
    \label{alg:unsupervised_seg}
\end{algorithmic}
\end{algorithm*}

\begin{algorithm*}
\begin{algorithmic}[1]
\Require Image $I$ (of size $(C,H,W)$), sparse user label mask $L$ of size $(H,W)$, desired featureset $FS$ (one of `Classical', `Upsampled ViT', `Hybrid'), a classifier $f$ - can be linear, logistic, random forest, XGB, \textit{etc.}
\Statex
    \If {$FS=\text{`Classical'}$}
        \State $F$, shape $(N_c,H,W)$ $\leftarrow$ Classical Weka-style features (edges, Gaussian blurs, ...)
    \ElsIf {$FS=\text{`Upsampled ViT'}$}
        \State $F$, shape $(k,H,W)$ $\leftarrow$ Get upsampled ViT features from Algorithm \ref{alg:upsampling}
    \ElsIf {$FS=\text{`Hybrid'}$}
        \State $F_c$, shape $(N_c,H,W)$ $\leftarrow$ Classical Weka-style features (edges, Gaussian blurs, ...)
        \State $F_v$, shape $(k,H,W)$ $\leftarrow$ Get upsampled ViT features from Algorithm \ref{alg:upsampling}
        \State $F$, shape $(N_c + k,H,W)$  $\leftarrow$ concat $F_c$ and $F_v$ in the channel dimension
    \EndIf
    
    \State $(X, Y)$ $\leftarrow$ get features of all labelled pixels $X$ from $F$ and their class values $Y$ from $L$; optionally sample; shuffle
    \State $f$ $\leftarrow$ fit classifier $f$ over training data $(X,Y)$
    \State $S$, shape $(H,W)$ $\leftarrow$ apply trained classifier $f$ to features of unlabelled pixels, predicting a class for each one
    
    \Return $S$, segmentation map of $I$ given user class labels in $L$ and desired featureset $FS$ and $f$, the trained classifier for re-use. 
    \caption{{\footnotesize \textcolor{black}{Weakly supervised segmentation with upsampled ViT features}}}
    \label{alg:weakly_supervised_seg}
\end{algorithmic}
\end{algorithm*}

\section{Further resolution comparisons}
\label{sup:more_qual_compare}

\begin{figure}[H]
\centering
    \includegraphics[width=1\linewidth]{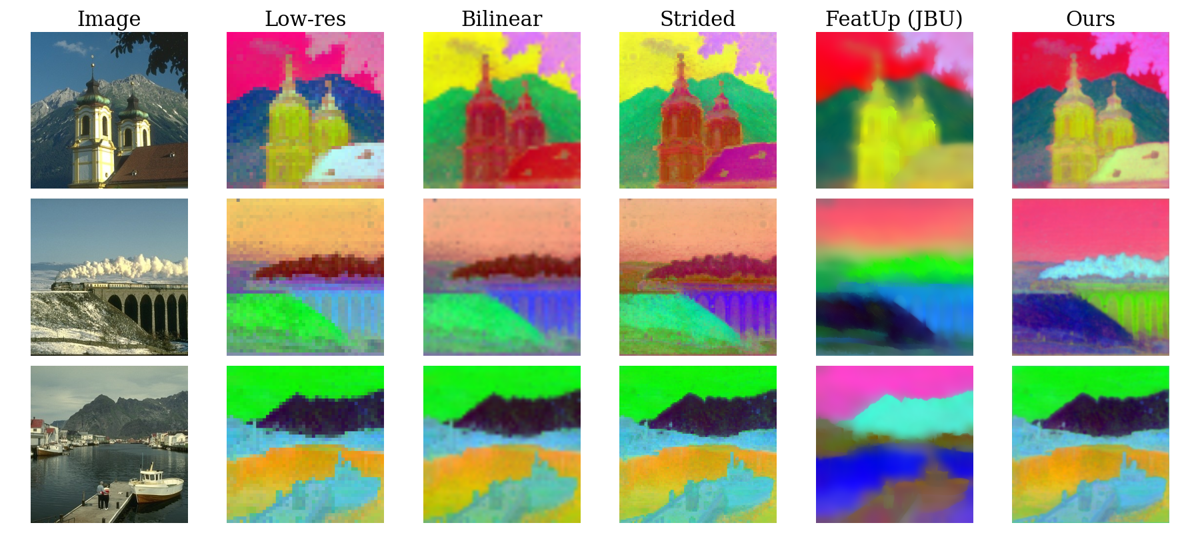}
    \caption{Cross-method comparison of feature upsampling methods on square crops of images from the BSD300 dataset. The model used for the all upsamplings except `FeatUp (JBU)' was DINO-S-8 - `FeatUp (JBU)' did not have a pretrained checkpoint for this resolution, so DINO-S-16 was used instead. The input image size was (384, 384).  }
    \label{fig:supp_method_compare}
\end{figure}

\textcolor{black}{Figure \ref{fig:supp_method_compare} shows a comparison of feature maps across different upsampling methods for a set of natural images.}

\begin{figure}[H]
\centering
    \includegraphics[width=1\linewidth]{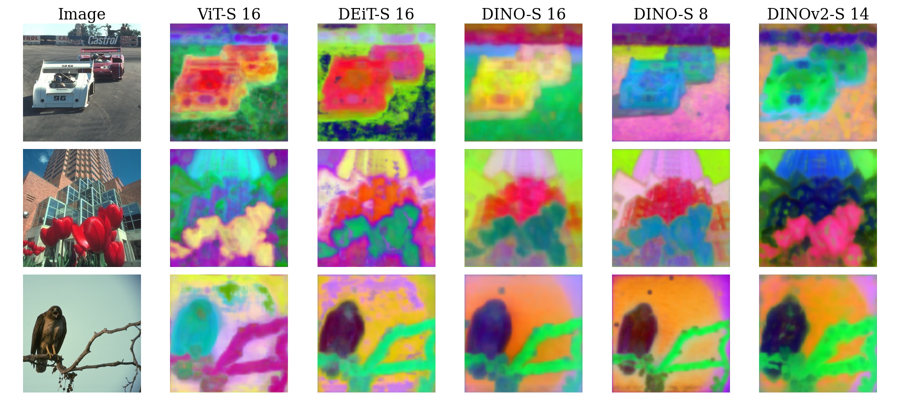}
    \caption{Cross-model comparison of our upsampling method on square crops of images from the BSD300 dataset with shift and flip transforms applied and a stride of 4. The input image size was (224, 224), the training resolution of ViT-S and DEiT-S. Note the DINO models can be applied to arbitrary resolutions, and improve as the image size increases. DINOv2 suffers the worst blurring, possibly because its patch size of 14 is not a multiple of the stride. }
    \label{fig:supp_model_compare}
\end{figure}

\textcolor{black}{Figure \ref{fig:supp_model_compare} shows a comparison of feature maps across different ViT backbones using our upsampling method (stride 4, shifts 1,2, flips). We see that models with a smaller patch size (DINO-S-8) achieve higher resolution, and that  DINOv2-S-14 suffers the worst blurring, moreso than models with a larger patch size (DINO-S-16, DEiT-s 16, ViT-s 16).}

\section{Small object retrieval and keypoint matching}
\begin{figure}[H]
\centering
    \includegraphics[width=1\linewidth]{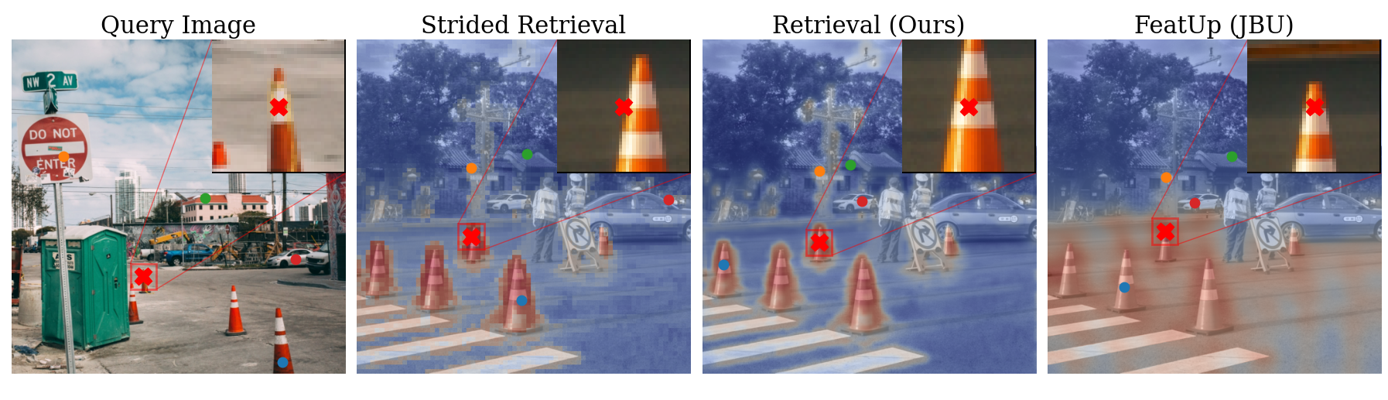}
    \caption{Small object retrieval (red cross) with \textcolor{black}{additional keypoint matching (blue, orange, green \& red coloured dots)} for various feature upsampling methods with DINOv2-S-14, stride 7. Following FeatUp, we compute high-resolution features for a query \textcolor{black}{(left, stop sign)} and target image \textcolor{black}{(right)}, then find the most similar point in the target image to a given query point (red cross) and additional keypoints). The colourmap indicates similarity to the query point, red for similarity, blue for dissimilarity. \textcolor{black}{The coloured dots are simply additional keypoints applied to the query image - we look for the most similar vector in the target image across the different upsampling methods}. We note our similarities are more localised than FeatUp's (JBU).   }
    \label{fig:retrieval}
\end{figure}

\begin{figure}[H]
\centering
    \includegraphics[width=1\linewidth]{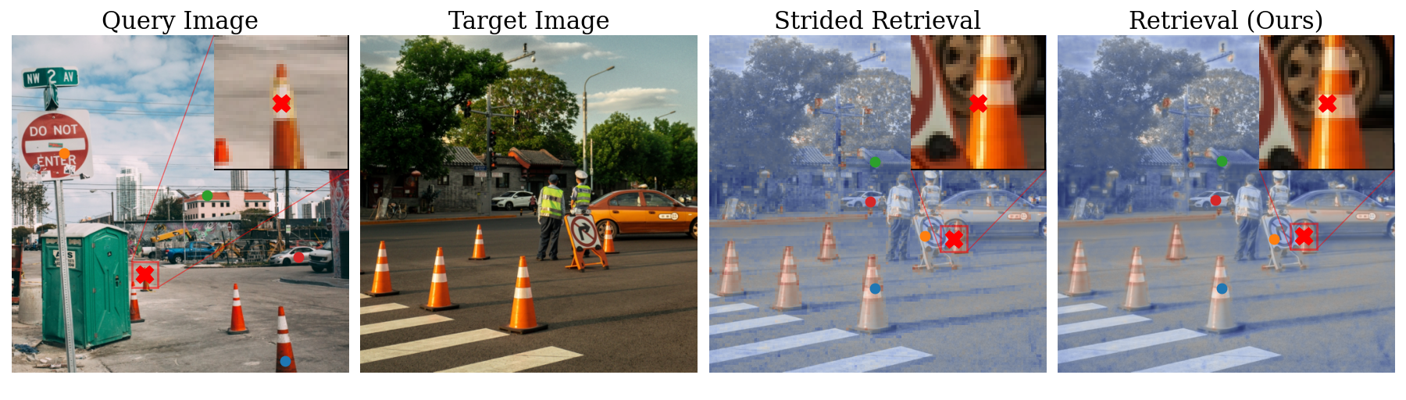}
    \caption{Small object retrieval and other keypoint matching with DINO-S-8 with stride 4. The similarity maps for the the query point (cross) are sharper than with DINOv2-S-14 (Figure \ref{fig:retrieval}), but emphasise colour information more \textit{i.e,} the cone is similar to the orange kerb and car and the stop signpost keypoint is matched to the red and white road sign rather than the background signpost. Using our approach gives a slightly more accurate retrieval than just using strided features.}
    \label{fig:supp_retrieval}
\end{figure}

\textcolor{black}{Following FeatUp \cite{FEATUP}, we perform keypoint matching and similarity visualisation using the upsampled features in Figures \ref{fig:retrieval} and \ref{fig:supp_retrieval}. In it, the feature vectors of series of keypoints on a query image (red cross, coloured dots) are taken, and the most similar points (in terms of cosine similarity) are found in a target image. We also plot the cosine similarity of every pixel to the feature vector red cross as a heatmap, red for similar, blue for dissimilar. This gives a measure of how localised the similarities are across the feature map in the target image. We see that the similarities are more localised with our method than Featup (JBU).  }

\section{Failure modes of the unsupervised workflow }
\begin{center}
    \begin{figure}[H]
        \includegraphics[width=\linewidth]{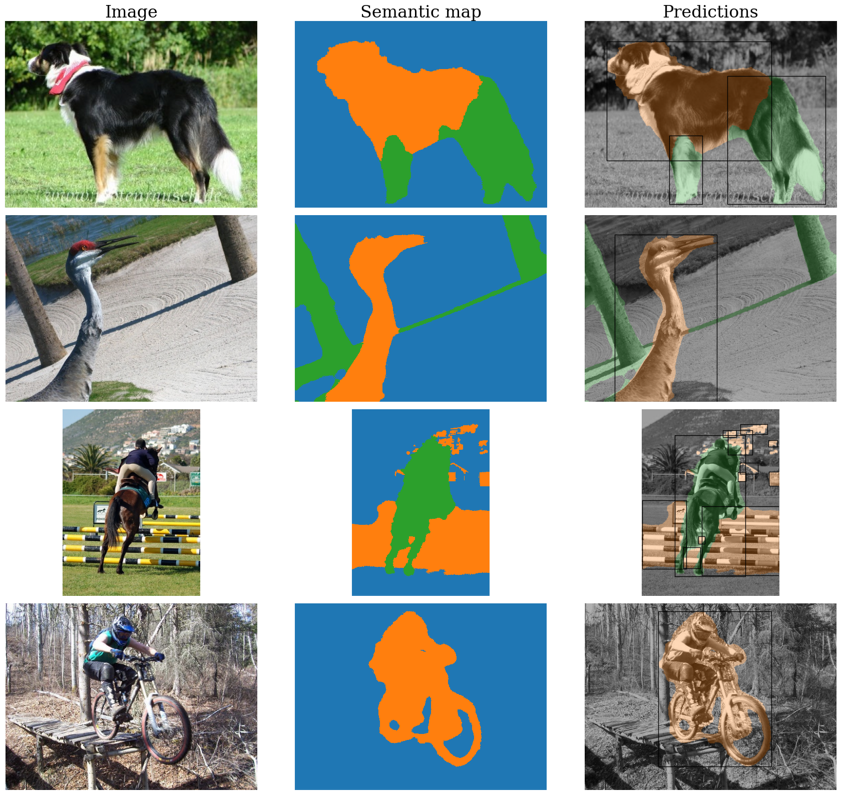}
        \caption{Examples of failure modes in the unsupervised segmentation workflow. First row: over-decomposition of foreground object into legs and body - this can be improved by increasing the merging threshold distance, but can also merge more unrelated classes. Second row: merging of unrelated classes \textit{i.e,} the shadow of an object and the object. Third and fourth row: over-merging objects close in space \textit{i.e,} the rider and the horse/bike.}
        \label{fig:supp_unsupervised_problems}
    \end{figure}
\end{center}

\textcolor{black}{We present failure modes of the unsupervised workflow in Figure \ref{fig:supp_unsupervised_problems}. The first is overdecomposing objects (i.e the dog's body and leges), the second is merging an object with its shadow and the last two are examples of over-merging (the horse/bike and its rider). These are, to some degree, functions of the relativity of the DINO features and the positional biases. Objects close in space to one another will be similar in terms of their positional information and foregroundedness, so their cosine distance will be smaller (and therefore be more likely to be merged). }

\section{Unsupervised segmentation examples}

\begin{center}
    \begin{figure}[H]
    \centering
        \includegraphics[width=0.8\linewidth]{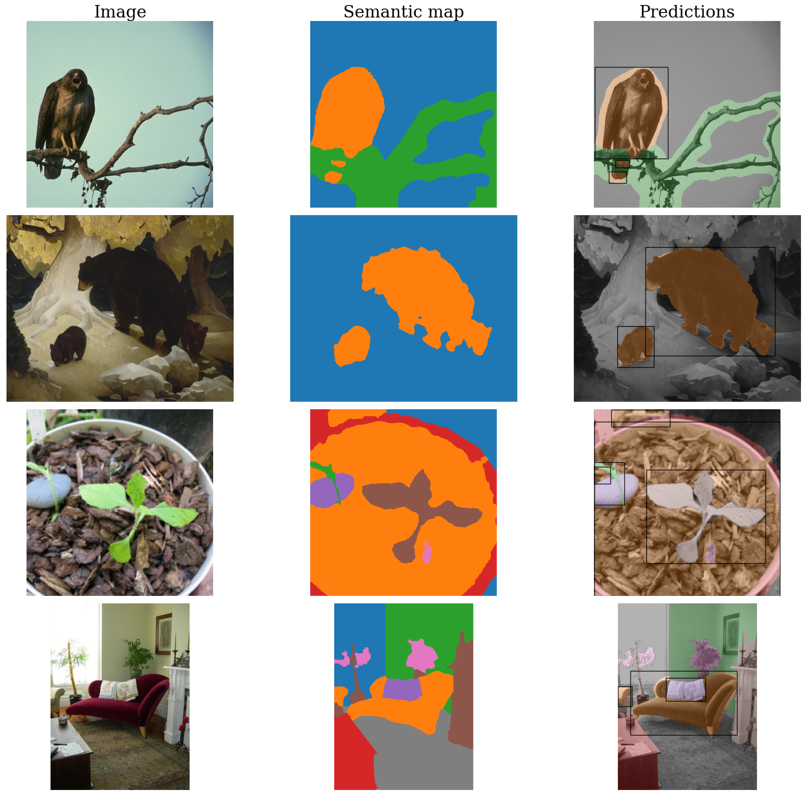}
        \caption{Example predictions of foreground objects and their segmentation using the unsupervised workflow. The background class (\textit{i.e,} the one with the lowest attention density) is show in blue. Note the blurring in the first image - this is more frequent in uncluttered scenes and worse at lower resolutions.}
        \label{fig:supp_unsupervised_examples}
    \end{figure}
\end{center}

\textcolor{black}{We show some examples of the unsupervised segmentation in Figure \ref{fig:supp_unsupervised_examples}, which demonstrates both the semantic segmentation from the determined semantic distance, and bounding boxes around the foreground classes based on the the attention density. The approach is able to separate touching objects of different classes (eagle, branch) and merge separate instances of the same class (couch, leaves, bears). These examples were produced using DINOv2-S-14, stride 4, shifts of 1 and 2px and flip transforms. }

\section{Weakly supervised segmentation}
\label{sec:supp_wss}

\subsection{T-cell case study}
\begin{center}
\begin{figure}[H]
\centering
    \includegraphics[width=0.8\linewidth]{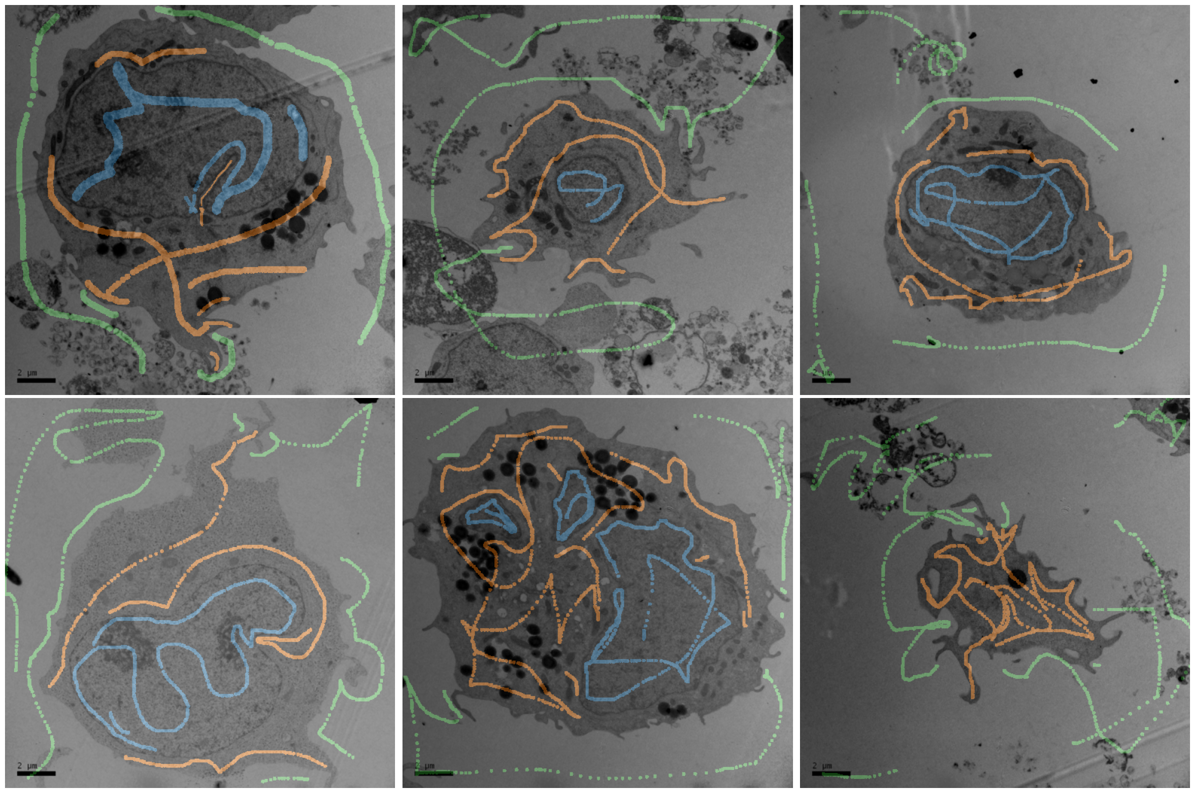}
    \caption{The labels and cells used to train each classifiers in the case study. Classfiers are trained to map from image features (be they classical, from a high-resolution ViT or a combination thereof) to the user-drawn labels, then applied to the unlabelled feature vectors, both of the cells here and the rest of the 129 unlabelled cells in the dataset. Note all cells (train + test) were resized from (1024, 1024) to (518, 518) during training and segmentation.}
    \label{fig:supp_wss_labels}
\end{figure}
\end{center}

\textcolor{black}{We show the labels we used to train the classifiers for Section \ref{sec:wss_t_cells} in Figure \ref{fig:supp_wss_labels}. We chose cells that covered the range of variation in the dataset (size, lack of nucleus, exposure, presence of multiple cells). We found it useful to label across the space, and to focus on boundaries between the classes.} 

\begin{center}
    \begin{figure}[H]
        \includegraphics[width=0.9\linewidth]{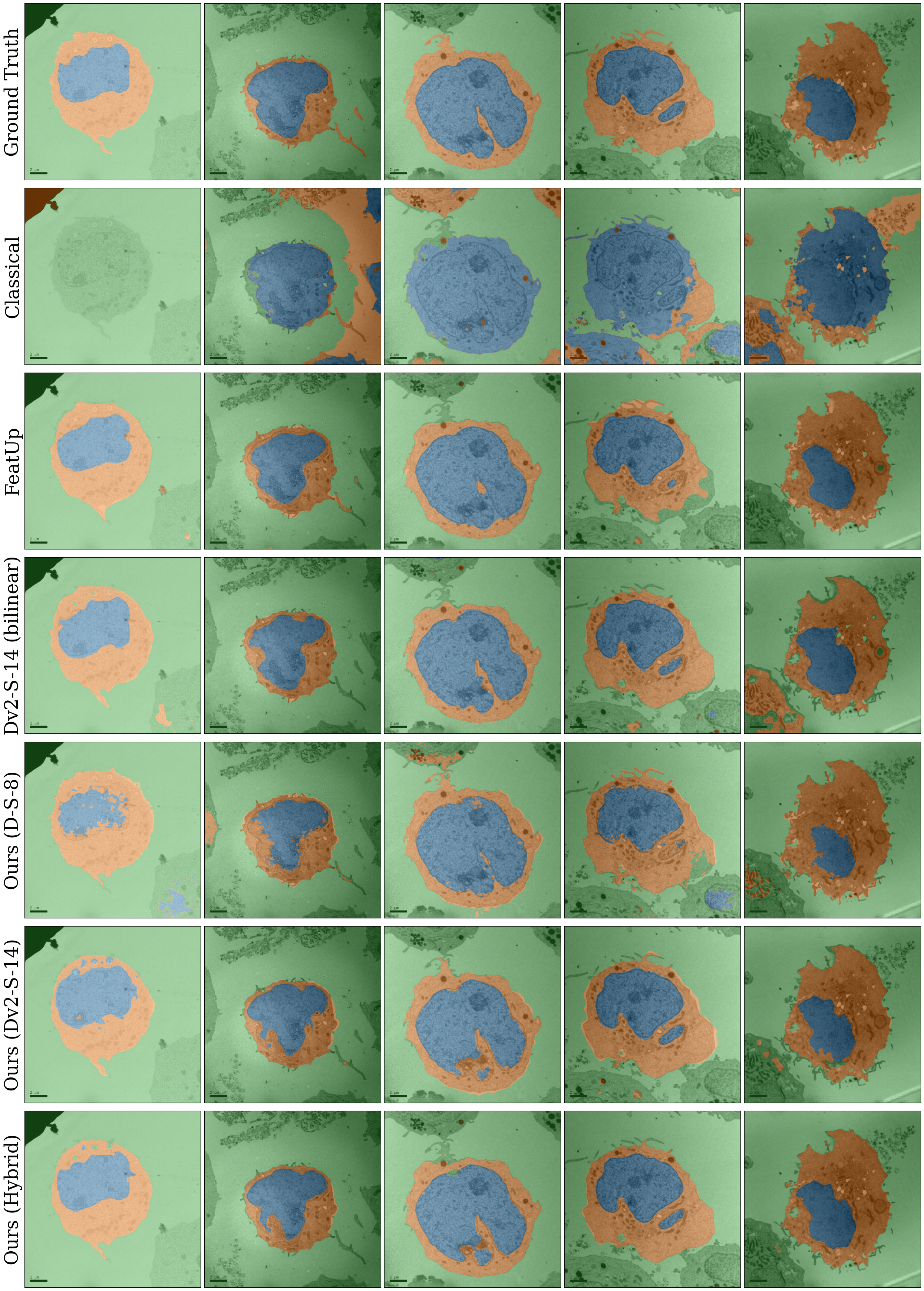}
        \caption{Example segmentations from applying classifiers trained on various feature-sets to unlabelled examples in the T-cell dataset. We note the upsampled ViT features produce segmentations that much more closely align with the ground truth, able to ignore other cluttered cells, handle the presence of multiple nuclei and varying exposure. These predictions have had a CRF applied. \textcolor{black}{We have abbreviated `DINO' to `D' \textit{i.e,} `D-S-8' is `DINO-S-8'.}}
        \label{fig:supp_wss_test}
    \end{figure}
\end{center}

\begin{center}
    \begin{figure}[H]
        \includegraphics[width=0.9\linewidth]{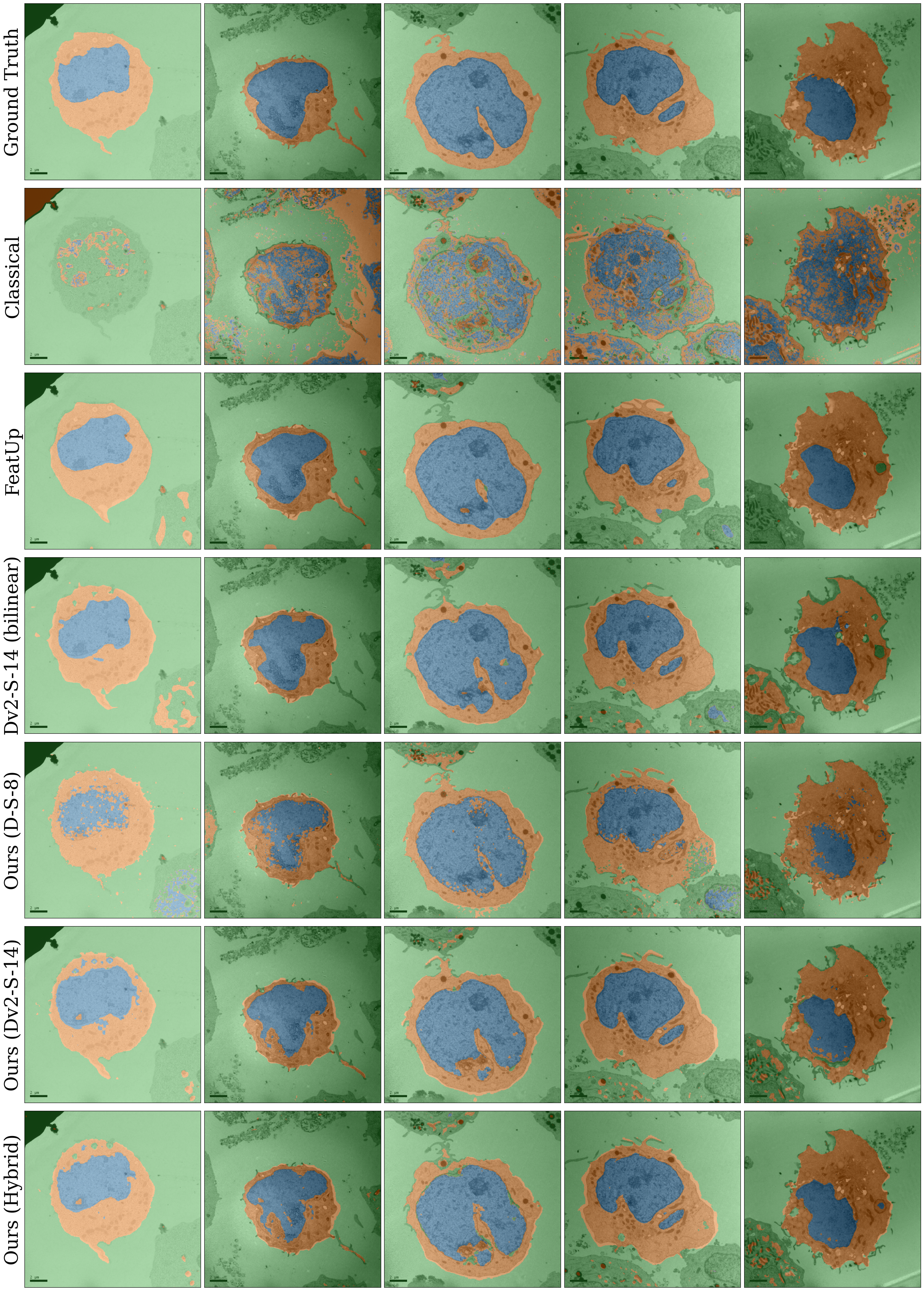}
        \caption{The same predictions as in Figure \ref{fig:supp_wss_test} without a CRF applied. \textcolor{black}{Again we have abbreviated `DINO' to `D' \textit{i.e,} `D-S-8' is `DINO-S-8'.}}
        \label{fig:supp_wss_test_nocrf}
    \end{figure}
\end{center}

\textcolor{black}{Figures \ref{fig:supp_wss_test} and Figures \ref{fig:supp_wss_test_nocrf} show predictions of the classifier trained with the labels in Figure \ref{fig:supp_wss_labels} on unlabelled cells in the dataset. We see greatly improved performance when using the upsampled ViT features relative to the classical features, regardless of whether CRF post-processing was used (Figure \ref{fig:supp_wss_test}) or not (Figure \ref{fig:supp_wss_test_nocrf}). }

\textcolor{black}{We include FeatUp and bilinear upsampling. Bilinear upsampling performs well, especially relative to its cost. However, it struggles to capture the fine features on the edges of cells (\textit{i.e,} which are smaller than the patch size of 14px), which is expected. These cells are large, so capturing fine detail is not necessary to achieve a high mIoU (in Table \ref{tab:wss}) However, in other materials, small features are highly important, like inclusions in metals or cracks in batteries, and the coarseness of the bilinear upsampling may then become a larger problem - we see this in Figure \ref{fig:supp_upsampler_ablation}.}

\subsection{Upsampler ablation}
\begin{center}
    \begin{figure}[H]
        \includegraphics[width=0.9\linewidth]{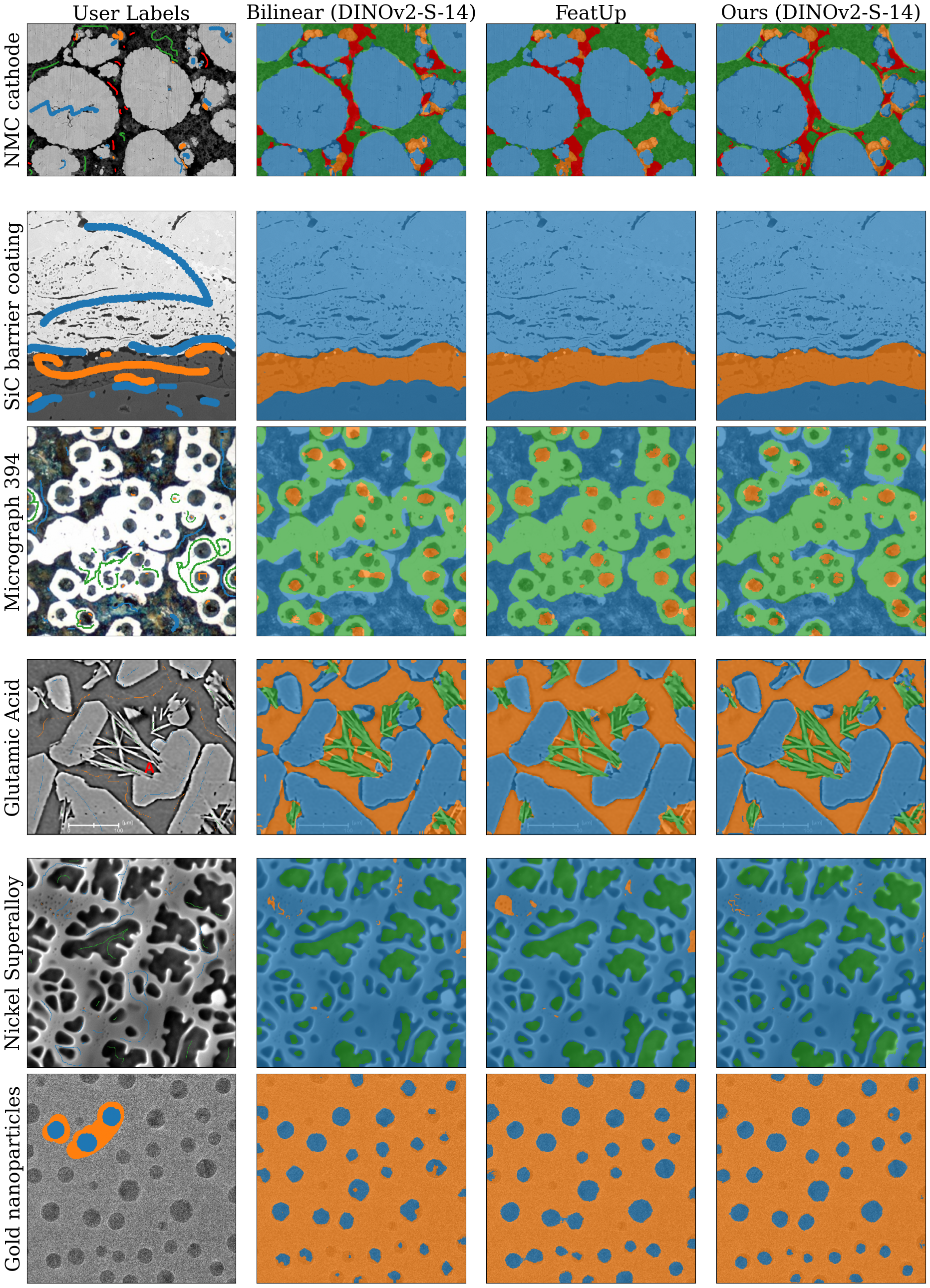}
        \caption{\textcolor{black}{Weakly supervised micrograph segmentation using different upsampled DINOv2 featuresets: bilinear upsampling, FeatUp, and our approach (stride 4, shifts 1,2 and flip transforms). We see that bilinear upsampling causes artefacts, and FeatUp, whilst better, displays some blurring (\textit{i.e,} between particles).}}
        \label{fig:supp_upsampler_ablation}
    \end{figure}
\end{center}

\textcolor{black}{Figure \ref{fig:supp_upsampler_ablation} shows weakly supervised segmentation of the micrographs in Figure \ref{fig:wss_examples} using different upsampling techniques. Bilinear upsampling suffers from clear artefacting, FeatUp and our approach are better, but still display some blurring. This motivates the hybrid approach, (Figure \ref{fig:wss_examples}), which combines the classical and upsampled ViT features to achieve a good balance of semantic and colour information.}

\begin{center}
    \begin{figure}[H]
    \centering
        \includegraphics[width=0.65\linewidth]{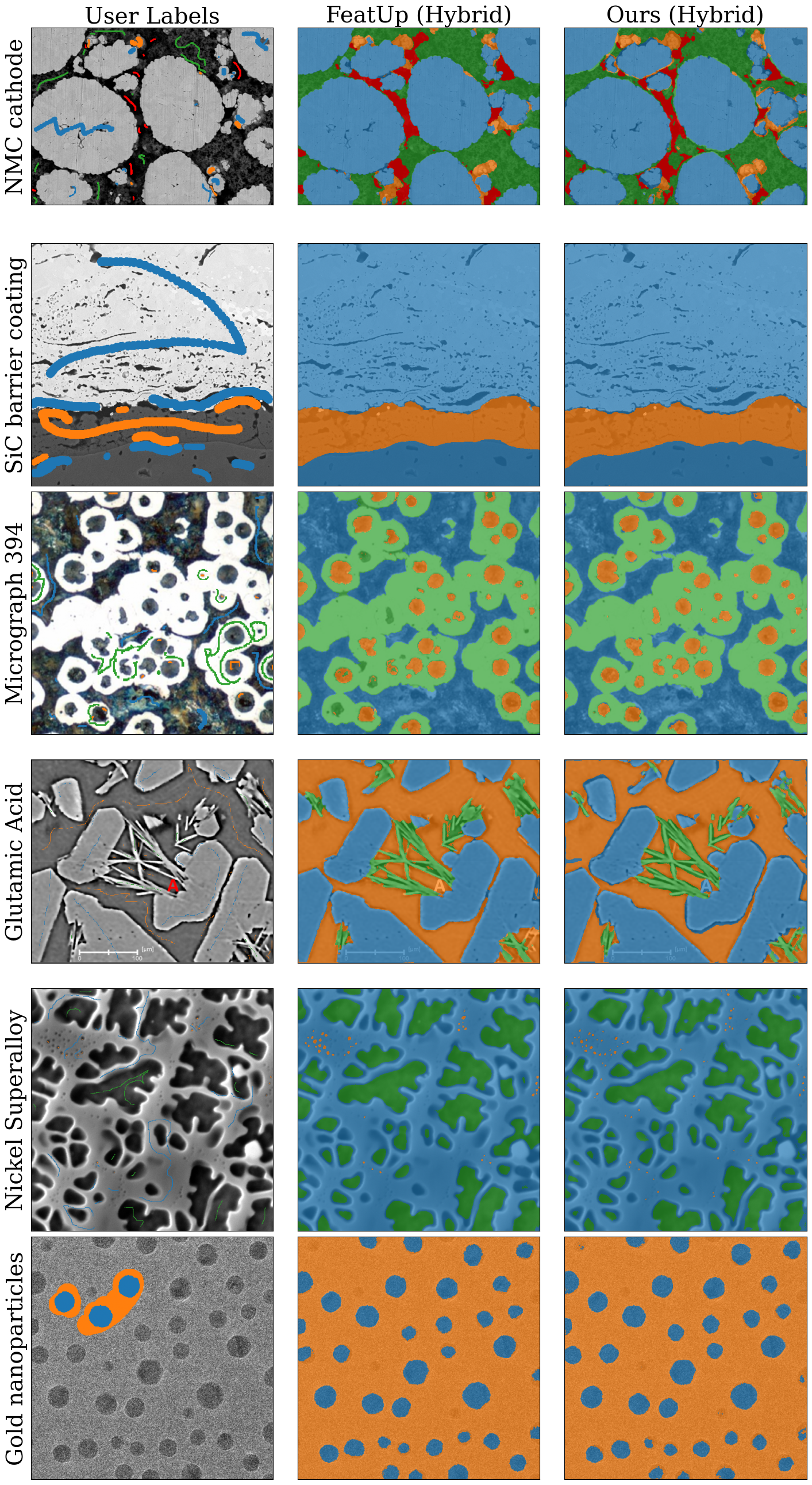}
        \caption{\textcolor{black}{Weakly supervised micrograph segmentation using different deep feature upsamplers in the `hybrid' scheme (\textit{i.e,} FeatUp deep features + classical or DINOv2-S-14 stride 4, shift 1,2 + classical). We see that our approach improves segmentation of fine features, like some spherodised graphite in Micrograph 394, parts of the secondary and tertiary precipitates in the Nickel superalloy.}}
        \label{fig:supp_hybrid_ablation}
    \end{figure}
\end{center}

\subsection{Classifier ablation}

\begin{center}
    \begin{figure}[H]
        \includegraphics[width=0.9\linewidth]{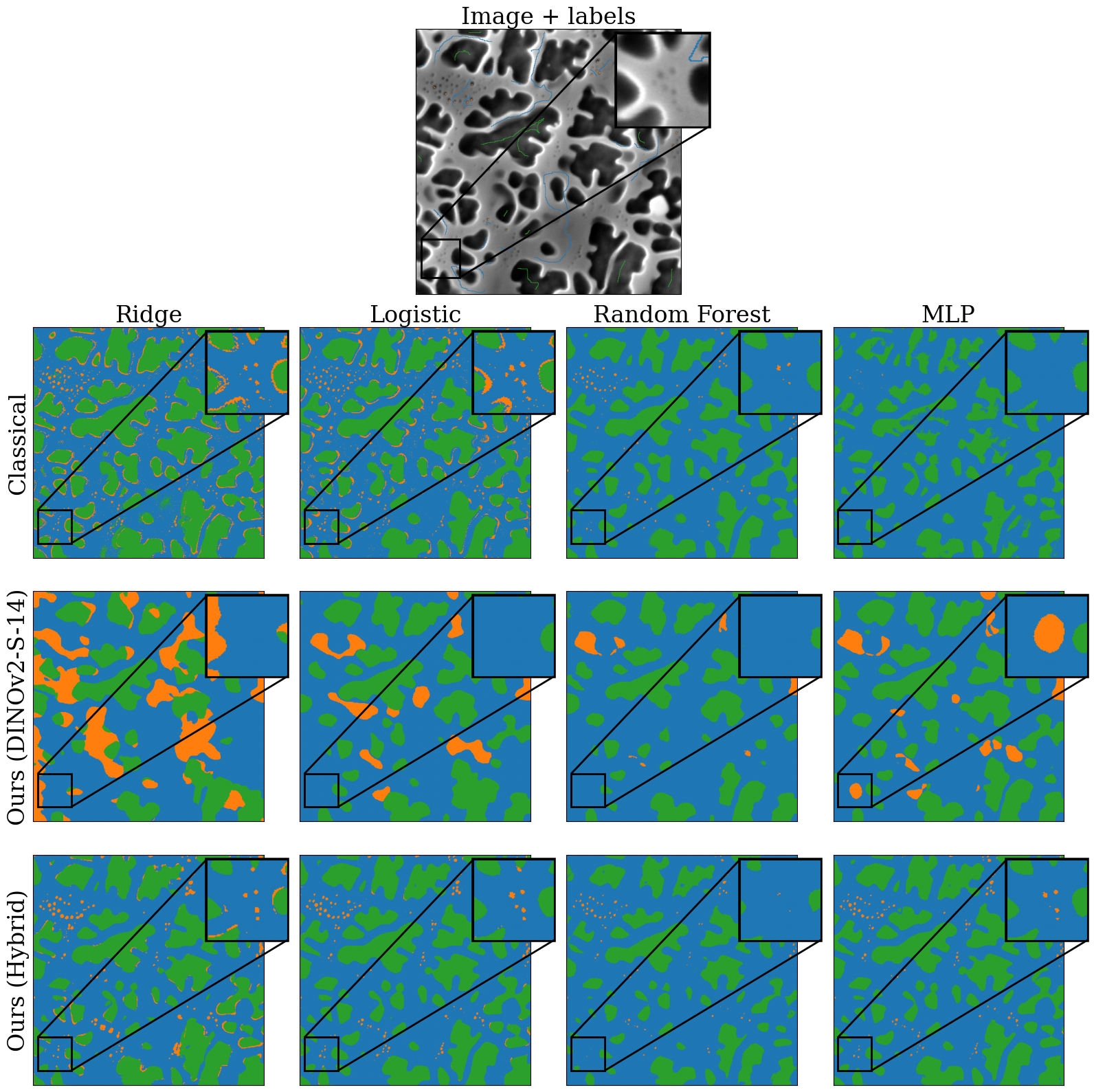}
        \caption{\textcolor{black}{Segmentations of a Nickel superalloy \cite{SUPERALLOY} for a fixed set of sparse labels as a function of features and classifier. We see the choice of classifier has little impact on the performance of the hybrid (classical + upsampled DINOv2-S-14) features, but does affect the classical features. }}
        \label{fig:supp_classifier_ablation}
    \end{figure}
\end{center}

\textcolor{black}{Figure \ref{fig:supp_classifier_ablation} shows the result of changing the classifier used for weakly supervised segmentation for different features. When using linear models (Ridge regression, logistic regression) the classical features struggle to differentiate the edges of the primary precipitates from the tertiary precipitates. This is because the classical features are linear, and cannot model non-linear relationships - Weka uses a (non-linear) random forest classifier to overcome this\cite{WEKA}.}

\textcolor{black}{The upsampled DINOv2-S-14 (stride 4, shift 1, 2, flips) features are not sufficiently high resolution to distinguish the precipitates regardless of classifier, though increasing classifier complexity (\textit{i.e,} using an MLP) does lead to an improved segmentation of the primary vs secondary precipitates.}

\textcolor{black}{Adding the classical features in the `Hybrid' scheme allows a good segmentation of each phase, which is not affected by classifier choice, even with linear models. Interestingly, the `Hybrid' scheme seems to perform better (\textit{i.e,} capture more tertiary precipitates) when using logistic regression or an MLP - we attribute this to smoother decision boundaries. The linear classifiers used default parameters, with 3000 iterations where applicable. The Random Forest was `Weka-style', with 200 trees, 2 features per node and unlimited depth. The MLP had three hidden layers of size 100, ReLU activations and Adam optimizer. }

\section{Transformations}
\label{sec:supp_transforms}

\begin{figure}[H]
    \includegraphics[width=\linewidth]{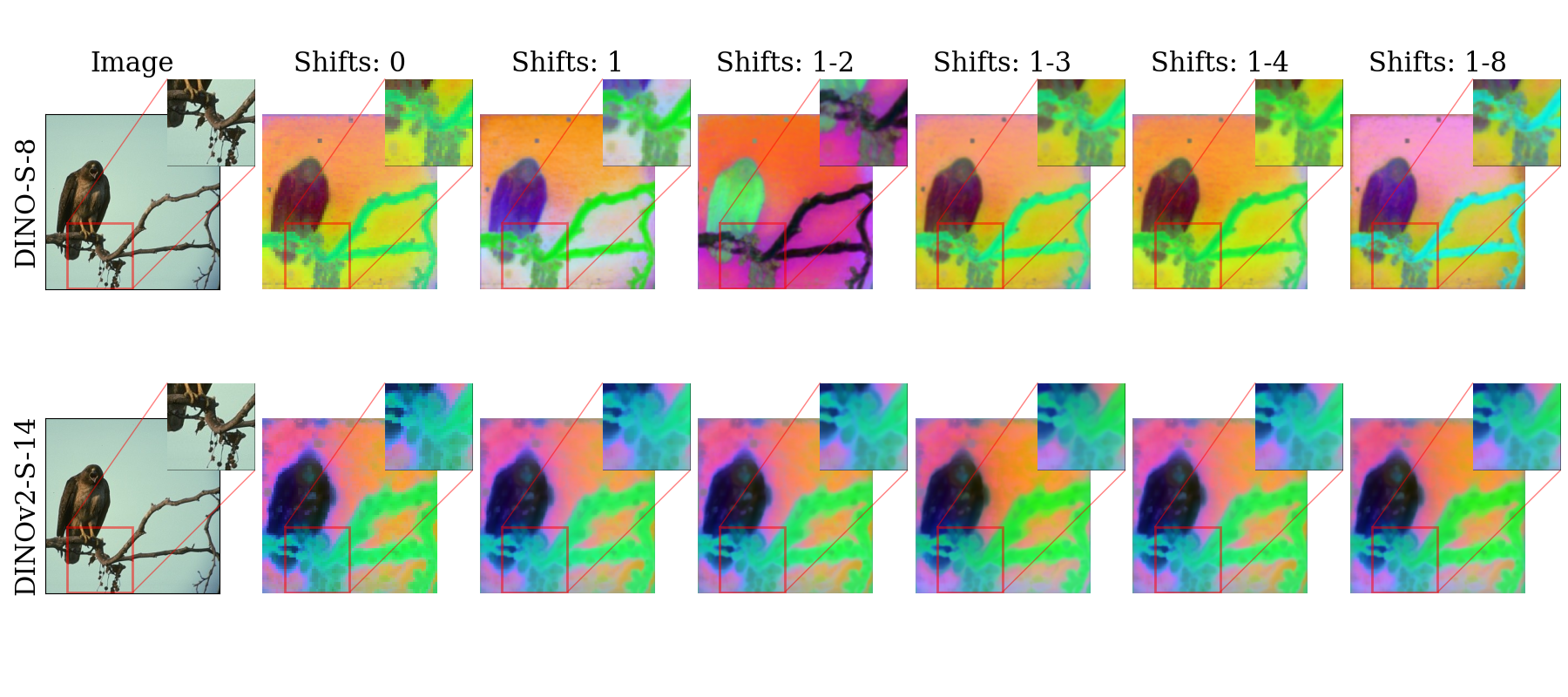}
    \caption{Visualisation of the impact of adding more shift transforms on the feature resolution for different ViTs with stride 4 - we note past shifts 1, 2 (\textit{i.e,} stride / 2) we see little, if any, improvement. }
    \label{fig:supp_shift_comparison}
\end{figure}

Based on experiments and shown in Figure \ref{fig:supp_shift_comparison}, for a ViT model with stride $S$, we found the best range of distances for the shift transforms to be $[1, S/2]$. When combined with the flip transforms this gives $4 \text{ (flips)} \times 8 \text{ (directions)} \times S / 2 = 32S \text{ transformations}$. Usually $S=4$ so the number of transforms was 64. This could be reduced by using a 4-neighbourhood rather than an 8-neighbourhood, at the cost of slightly reduced performance. Although 4 flip transforms were used (no flip, flip vertical, no flip, flip horizontal), three would also work (no flip, flip vertical, flip horizontal).

For the object detection and localization studies, a stride of 4 was used, as were the shifts [1, 2] as well as flip transforms. Flip transforms were useful to average out positional information to decrease the semantic distance between instances of the same class on different sides of the image.

For the cell segmentation case study, a stride of 4 was used as well as the shifts [1, 2]. No flip transforms were used, positional information was useful in this case to identify the central cell.

For the weakly supervised segmentation examples, a stride of 4 was used, as were the shifts [1, 2] as well as flip transforms (materials segmentation is a more homogenous problem).

\begin{table}[H]
\centering
    \begin{tabular}{l c c }
        \toprule
        \textbf{Feature-set} & \textbf{mIoU} \\
        \midrule
        DINO-S-8 (no shift) & 0.7446 \\
        DINOv2-S-14 (no shift) & 0.7915 \\
        \midrule
        DINO-S-8 (shift 1, 2) & 0.7764 \\
        DINOv2-S-14 (shift 1, 2) & 0.7974 \\
        \midrule
    \end{tabular}
    \caption{mIoU across the cell TEM dataset for ViT features (stride 4, no CRF) without and with shift transforms applied - the performance gain is modest for DINOv2 and more marked for DINO.}
    \label{tab:ablation_wss_shift}
\end{table}

\begin{table}[H]
\centering
    \begin{tabular}{l c c }
        \toprule
        \textbf{Feature-set} & \textbf{mIoU} \\
        \midrule
        DINOv2-S-14 (no transforms) & 0.638 \\
        DINOv2-S-14 (flip, no shift) & 0.6465 \\
        DINOv2-S-14 (no flip, shift) &  0.6477 \\
        \midrule
        DINOv2-S-14 (flips, shift 1, 2) & 0.654 \\
        \midrule
    \end{tabular}
    \caption{IoU on DUTs foreground object segmentation as a function of transforms applied}
    \label{tab:ablation_duts_tr}
\end{table}

\begin{table}[H]
\centering
    \begin{tabular}[t]{l c c c }
            \toprule
             \textbf{Method} & \textbf{VOC07} & \textbf{VOC12} \\
            \midrule
            DINO \cite{DINO, VIT_UNSUPERVISED_OD_SURVEY} & 0.458 & 0.462  \\
            \textit{Ours} \textcolor{black}{(DINOv2-S-14, NU)} & 0.518 & 0.553  \\
            \textit{Ours} \textcolor{black}{(DINOv2-S-14, NU, multi)} & 0.614 & 0.617  \\
            \textit{Ours} \textcolor{black}{(DINOv2-S-14)} & \textbf{0.554 } & \textbf{0.572} \\
            \textit{Ours} \textcolor{black}{(DINOv2-S-14, multi)}  & \textbf{0.718 }& \textbf{0.725}   \\ 
            \midrule
                    
        \end{tabular}
\caption{\textcolor{black}{CorLoc on VOC07 \& VOC12 for our object localization framework. `NU' refers to `No Upsampling': no changes to model stride or pixel shift transforms and nearest-neighbour upsampling of the features. We see that adding our upsampling increases performance.  }}
        \label{tab:ablation_voc_}
\end{table}

\subsection{Impact of transformations on WSS predictions}
\label{sec:supp_transforms_wss}

\begin{center}
    \begin{figure}[H]
        \includegraphics[width=\linewidth]{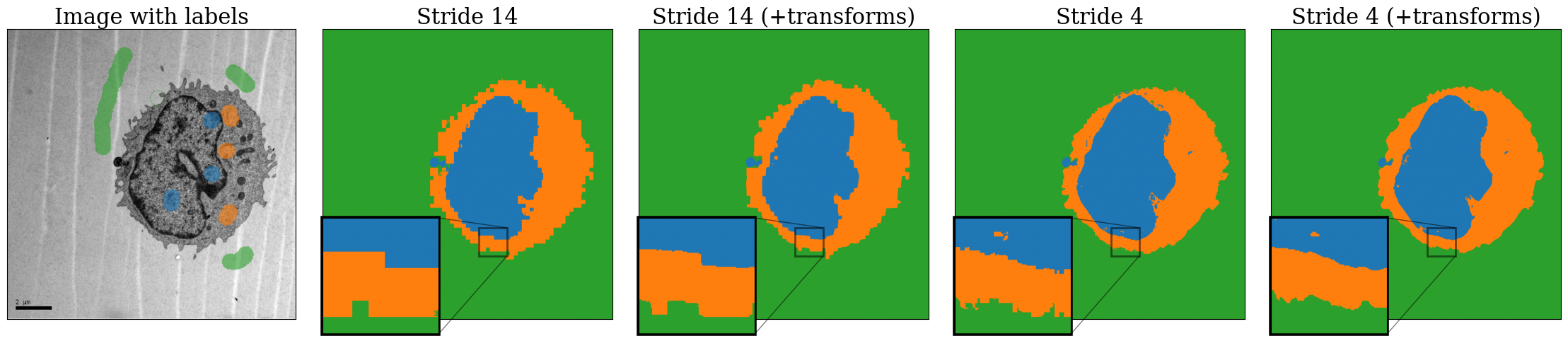}
        \caption{Impact of changing the model stride and adding shift transforms on the resulting weakly-supervised segmentation. Note classical features have been added for all examples, \textit{i.e,} this is the hybrid scheme.  At high strides (stride = model patch size), we get blocky predictions, which is improved as the stride reduces and transformations (shift + flips) are added.}
        \label{fig:supp_wss_transforms}
    \end{figure}
\end{center}

\subsection{Impact of transformations on time/memory cost}
\label{sec:supp_transforms_memory_cost}

\textcolor{black}{Each transformation (a single shift or flip) corresponds to one additional forward pass of the original ViT model (in most cases DINOv2). This means the peak time and memory cost scales linearly with the number of transforms. If they all fit in memory, this can be done in a single forward pass, so the time cost is constant. We call this approach `batched' and the opposite - computing the features of each transformation sequentially - `sequential'.}

\textcolor{black}{In order to compose two sets of transformations (like flip and pixel-shifts when averaging out positional bias), $T_{1}=\{flip_{LR}, flip_{UD}\}$ and $T_{2} =\{shift_{\uparrow, 1}, shift_{\rightarrow, 1}, ... \}$ the total number of transforms is the cartesian product, \textit{i.e} $N_{transforms}=T_1 \times T_2 = |T_1||T_2|$. Note that for each example there is also a forward pass of the untransformed image.  We present the peak time and memory cost with different transformations in Table \ref{tab:supp_transforms_cost} for DINOv2-S-14 and DINO-S-8 stride 4, noting that it aligns well the findings in Figure \ref{fig:supp_practical}.}

\begin{table}[H]
\centering
    \begin{tabular}{l c c c c}
        \toprule
        \textbf{Transformations} & \textbf{$N_{t}$} & \textbf{Memory (MB)} & \textbf{Time (s)} \\
        \midrule
        \textbf{Batched (DINOv2-S-14)} & &  &  \\
        No transforms & 0 & 281 & 0.04 & \\
        Von Neumann shift (1) & 4 & 506 & 0.12 & \\
        Moore shift (1) & 8 & 551 & 0.22 & \\
        Moore Shift (1, 2)  & 16 & 1002 & 0.42 & \\
        Moore Shift (1, 2) + flips  & 64 & 4008 &  1.57& \\
        \textbf{Sequential (DINOv2-S-14)} & & & \\
        No transforms & 0 & 281 & 0.04 & \\
        Von Neumann shift (1) & 4 & 464 & 0.14 & \\
        Moore shift (1) & 8 & 466 & 0.26 & \\
        Moore Shift (1, 2)  & 16 & 472 & 0.49 & \\
        Moore Shift (1, 2) + flips  & 32 & 508 & 1.98 & \\
        \textbf{Batched (DINO-S-8)} & &  &  \\
        No transforms & 0 & 282 & 0.03 & \\
        Von Neumann shift (1) & 4 & 507 & 0.13 & \\
        Moore shift (1) & 8 & 553 & 0.22 & \\
        Moore Shift (1, 2)  & 16 & 1025 & 0.40 & \\
        Moore Shift (1, 2) + flips  & 64 & 4100 &  1.55& \\
    \end{tabular}
    \caption{\textcolor{black}{Time and memory cost scaling as a function of added transformations for different DINO models with a stride of 4. }}
    \label{tab:supp_transforms_cost}

\end{table}

\end{document}